\pdfoutput=1

\documentclass[11pt]{article}

\usepackage[]{EMNLP2022}

\usepackage{times}
\usepackage{latexsym}
\usepackage{enumitem}

\usepackage[T1]{fontenc}

\usepackage[utf8]{inputenc}

\usepackage{microtype}

\usepackage{inconsolata}

\usepackage{graphicx}
\usepackage{xspace}
\usepackage{tabularx}
\usepackage{booktabs}
\usepackage{multirow}
\usepackage{comment}
\usepackage{caption}
\usepackage{subcaption}
\usepackage{textgreek}
\usepackage{amssymb}
\usepackage{amsmath}
\usepackage{color}
\usepackage{color, colortbl}
\usepackage{newunicodechar}
\usepackage{tipa}
\usepackage[T4,T1]{fontenc}
\usepackage{siunitx}
\usepackage{pifont} 

\newcount\Comments  
\usepackage{color}
\newcommand{\kibitz}[2]{\ifnum\Comments=1\textcolor{#1}{#2}\fi}

\newenvironment{tfour}{\fontencoding{T4}\selectfont}{}

\newcommand{\cmark}{\ding{51}}%
\newcommand{\xmark}{\ding{55}}%
\newunicodechar{ɔ}{\fon}
\newunicodechar{ɖ}{\fon}
\definecolor{Color}{gray}{0.8}

\newcommand*{\yoruba}{Yor\`ub\'a\xspace}
\newcommand*{\ghomala}{Ghom\'al\'a'\xspace}
\newcommand*{\ewe}{\'Ew\'e\xspace}
\newcommand*{\zulu}{isiZulu\xspace}
\newcommand*{\xhosa}{isiXhosa\xspace}
\newcommand*{\shona}{chiShona\xspace}
\newcommand*{\swahili}{Kiswahili\xspace}
\newcommand*{\mafand}{MAFAND-MT\xspace}

\title{MasakhaNER 2.0: Africa-centric Transfer Learning for Named Entity Recognition}

\author{\normalsize David Ifeoluwa Adelani$^{1,2,*}$, Graham Neubig$^{3}$, Sebastian Ruder$^{4}$, Shruti Rijhwani$^{3}$,  \\
\textbf{\normalsize Michael Beukman$^{5*}$, Chester Palen-Michel$^{6*}$, Constantine Lignos$^{6*}$, Jesujoba O. Alabi$^{1*}$, } \\
\textbf{ \normalsize Shamsuddeen H. Muhammad$^{7*}$, Peter Nabende$^{8*}$, Cheikh M. Bamba Dione$^{9*}$, Andiswa Bukula$^{10}$, }\\
\textbf{\normalsize Rooweither Mabuya$^{10}$, Bonaventure F. P. Dossou$^{11*}$, Blessing Sibanda$^{*}$, Happy Buzaaba$^{12*}$, } \\
\textbf{\normalsize Jonathan Mukiibi$^{8*}$, Godson Kalipe$^{*}$, Derguene Mbaye$^{13*}$, Amelia Taylor$^{14*}$, Fatoumata Kabore$^{15*}$}, \\
\textbf{\normalsize Chris Chinenye Emezue$^{16*}$, Anuoluwapo Aremu$^{*}$, Perez Ogayo$^{3*}$, Catherine Gitau$^{*}$}, \\
\textbf{\normalsize Edwin Munkoh-Buabeng$^{17*}$, Victoire M. Koagne$^{*}$, Allahsera Auguste Tapo$^{18*}$, Tebogo Macucwa$^{19*}$},  \\
\textbf{\normalsize Vukosi Marivate$^{19*}$, Elvis Mboning$^{*}$, Tajuddeen Gwadabe$^{*}$,  Tosin Adewumi$^{20*}$, } \\
\textbf{\normalsize  Orevaoghene Ahia$^{21*}$, Joyce Nakatumba-Nabende$^{8*}$, Neo L. Mokono$^{19*}$, Ignatius Ezeani$^{22*}$,} \\ 
\textbf{\normalsize  Chiamaka Chukwuneke$^{22*}$,  Mofetoluwa Adeyemi$^{23*}$, Gilles Q. Hacheme$^{24*}$, Idris Abdulmumin$^{25*}$, } \\ 
\textbf{\normalsize Odunayo Ogundepo$^{23*}$, Oreen Yousuf$^{15*}$, Tatiana Moteu Ngoli$^{*}$, Dietrich Klakow$^{1}$ } \\\\
\footnotesize
$^*$Masakhane NLP, $^1$Saarland University, Germany, $^2$University College London, UK, $^3$Carnegie Mellon University, USA,  \\
\footnotesize
$^4$Google Research,
$^5$University of the Witwatersrand, South Africa,
$^6$ Brandeis University, USA, 
$^{7}$LIAAD-INESC TEC, Portugal,
\\
\footnotesize
$^{8}$Makerere University, Uganda
$^{9}$University of Bergen, Norway,
$^{10}$SADiLaR, South Africa,
$^{11}$Mila Quebec AI Institute, Canada, \\
\footnotesize
$^{12}$RIKEN Center for AI Project, Japan, 
$^{13}$Baamtu, Senegal,
$^{14}$Malawi University of Business and Applied Science, Malawi, \\
\footnotesize
$^{15}$Uppsala University, Sweden,
$^{16}$TU Munich, Germany,
$^{17}$TU Clausthal, Germany, 
$^{18}$Rochester Institute of Technology, USA, \\
\footnotesize
$^{19}$University of Pretoria, South Africa,
$^{20}$Luleå University of Technology, Sweden,
$^{21}$University of Washington, USA,\\
\footnotesize
$^{22}$Lancaster University, UK, 
$^{23}$University of Waterloo, Canada,
$^{24}$Ai4innov, France, 
$^{25}$Ahmadu Bello University, Nigeria.
}

\begin{document}
\maketitle
\begin{abstract}



African languages are spoken by over a billion people, but are underrepresented in NLP research and development. The challenges impeding progress include the limited availability of annotated datasets, as well as a lack of understanding of the settings where current methods are effective. In this paper, we make progress towards solutions for these challenges, focusing on the task of named entity recognition (NER). We create the largest human-annotated NER dataset for 20 African languages, and we study the behavior of state-of-the-art cross-lingual transfer methods in an Africa-centric setting, demonstrating that the choice of source language significantly affects performance. We show that choosing the best transfer language improves zero-shot F1 scores by an average of 14 points across 20 languages compared to using English. Our results highlight the need for benchmark datasets and models that cover typologically-diverse African languages. 


\end{abstract}

\section{Introduction}

Many African languages are spoken by millions or tens of millions of speakers.
However, these languages are poorly represented in NLP research, and the development of NLP systems for African languages is often limited by the lack of datasets for training and evaluation \cite{adelani-etal-2021-masakhaner}. 

Additionally, while there has been much recent work in using zero-shot cross-lingual transfer~\cite{ponti-etal-2020-xcopa,pfeiffer-etal-2020-mad,ebrahimi-etal-2022-americasnli} 
to improve performance on tasks for low-resource languages with multilingual pretrained language models (PLMs)~\cite{devlin-etal-2019-bert,conneau-etal-2020-unsupervised}, the settings under which contemporary transfer learning methods work best are still unclear~\cite{pruksachatkun-etal-2020-intermediate,lauscher-etal-2020-zero,xia-etal-2020-predicting}. 
For example, several methods use English as the source language because of the availability of training data across many tasks~\cite{pmlr-v119-hu20b,ruder-etal-2021-xtreme}, 
but there is evidence that English is often not the best transfer language~\citep{lin-etal-2019-choosing, de-vries-etal-2022-make, oladipo2022an}, and the process of choosing the best source language to transfer from remains an open question.

There has been recent progress in creating benchmark datasets for training and evaluating 
models in African languages for several tasks such as machine translation~\cite{nekoto_etal_2020_participatory,reid-etal-2021-afromt,adelani-etal-2021-effect,adelani_mafand,abdulmumin-etal-2022-hausa}, and sentiment analysis~\cite{yimam-etal-2020-exploring,Muhammad2022NaijaSentiAN}. 
In this paper, we focus on the standard NLP task of named entity recognition (NER) because of its utility in downstream applications such as question answering and information extraction. For NER, annotated datasets exist only in a few African languages~\cite{adelani-etal-2021-masakhaner,tigrinya_ner}, the largest of which is the MasakhaNER dataset~\cite{adelani-etal-2021-masakhaner} (which we call MasakhaNER 1.0 in the remainder of the paper).
While MasakhaNER 1.0 covers 10 African languages spoken mostly in West and East Africa, it does not include any  languages spoken in Southern Africa, which have distinct syntactic and morphological characteristics and are spoken by 40 million people. 

In this paper, we tackle two current challenges in developing NER models for African languages: (1) the lack of typologically- and geographically-diverse evaluation datasets for African languages; and (2) choosing the best transfer language for NER in an Africa-centric setting, which has not been previously explored in the literature.

To address the first challenge, we create the MasakhaNER 2.0 corpus, the largest human-annotated NER dataset for African languages. MasakhaNER 2.0 contains annotated text data from 20 languages widely spoken in Sub-Saharan Africa and is complementary to the languages present in previously existing datasets~\citep[e.g.,][]{adelani-etal-2021-masakhaner}. We discuss our annotation methodology as well as perform benchmarking experiments on our dataset with state-of-the-art NER models based on multilingual PLMs.

In addition, to better understand the effect of source language on transfer learning, we extensively analyze different features that contribute to cross-lingual transfer, including linguistic characteristics of the languages (i.e., typological, geographical, and phylogenetic features) as well as data-dependent features such as entity overlap across source and target languages~\citep{lin-etal-2019-choosing}. We demonstrate that choosing the best transfer language(s) in both single-source and co-training setups leads to large improvements in NER performance in zero-shot settings; our experiments show an average of a 14 point increase in F1 score as compared to using English as source language across 20 target African languages.  We release the data, code, and models on Github\footnote{\url{https://github.com/masakhane-io/masakhane-ner/tree/main/MasakhaNER2.0}} 

\section{Related Work}

\paragraph{African NER Datasets} 
There are some human-annotated NER datasets for African languages such as the SaDiLAR NER corpus~\cite{eiselen-2016-government} covering 10 South African languages, LORELEI ~\cite{strassel-tracey-2016-lorelei}, which covers nine African languages but is not open-sourced, and some individual language efforts for Amharic~\cite{Jibril2022ANECAA}
, 
\yoruba~\cite{alabi-etal-2020-massive}, Hausa~\cite{hedderich-etal-2020-transfer}, and Tigrinya~\cite{tigrinya_ner}. Closest to our work is the MasakhaNER 1.0 corpus~\cite{adelani-etal-2021-masakhaner}, which covers 10 widely spoken languages in the news domain, but excludes languages from the southern region of Africa like \zulu, \xhosa, and \shona with distinct syntactic features (e.g., noun prefixes and capitalization in between words) which limits transfer learning from other languages. We include five languages from Southern Africa in our new corpus. 

\paragraph{Cross-lingual Transfer}
Leveraging cross-lingual transfer has the potential to drastically improve model performance without requiring 
large amounts of data in the target language~\citep{conneau-etal-2020-unsupervised} but it is not always clear from which language we must transfer from~\citep{lin-etal-2019-choosing, de-vries-etal-2022-make}. To this end, recent work investigates methods for selecting good transfer languages and informative features.
For instance, token overlap between the source and target language is a useful predictor of transfer performance for some tasks~\citep{lin-etal-2019-choosing,wu-dredze-2019-beto}. Linguistic distance~\citep{lin-etal-2019-choosing,de-vries-etal-2022-make}, word order~\citep{k2020Crosslingual, pires-etal-2019-multilingual} and script differences~\citep{de-vries-etal-2022-make}, and syntactic similarity~\citep{karamolegkou-stymne-2021-investigation} have also been shown to impact performance.
Another research direction attempts to build models of transfer performance that predicts the best transfer language for a target language by using some linguistic and data-dependent features~\citep{lin-etal-2019-choosing,ahuja-etal-2022-multi}.

\section{Languages and Their Characteristics}
\begin{table*}[th!]
 \footnotesize
 \begin{center}
 \resizebox{\textwidth}{!}{%
  \begin{tabular}{lllr|llrr}
    \toprule
     & &\textbf{African} & \textbf{No. of}  &  \multicolumn{2}{c}{\textbf{}} & \textbf{\% Entities}   & \textbf{\#}\\
    \textbf{Language} & \textbf{Family} & \textbf{Region} & \textbf{Speakers}   & \textbf{Source} & \textbf{Train / dev / test}  & \textbf{in Tokens}   & \textbf{Tokens} \\
    \midrule
    Bambara (\texttt{bam}) & NC / Mande & West & 14M & \mafand~\cite{adelani_mafand} & 4462/ 638/ 1274 & 6.5 & 155,552\\
    \ghomala (\texttt{bbj}) & NC / Grassfields &Central& 1M & \mafand~\cite{adelani_mafand} & 3384/ 483/ 966 & 11.3 & 69,474 \\
    \ewe (\texttt{ewe}) & NC / Kwa &West& 7M  & \mafand~\cite{adelani_mafand} & 3505/ 501/ 1001 & 15.3 & 90420  \\
    Fon (\texttt{fon}) & NC / Volta-Niger &West& 2M & \mafand~\cite{adelani_mafand}  & 4343/ 621/ 1240  & 8.3 & 173,099\\
    Hausa (\texttt{hau}) & Afro-Asiatic / Chadic &West& 63M & Kano Focus and Freedom Radio & 5716/ 816/ 1633  & 14.0 & 221,086\\
    Igbo (\texttt{ibo}) & NC / Volta-Niger &West& 27M  & IgboRadio and Ka \d{O}d\d{I} Taa  & 7634/ 1090/ 2181  & 7.5 & 344,095\\
    Kinyarwanda (\texttt{kin}) & NC / Bantu &East& 10M & IGIHE, Rwanda & 7825/ 1118/ 2235  & 12.6 & 245,933 \\
    Luganda (\texttt{lug}) & NC / Bantu &East& 7M & \mafand~\cite{adelani_mafand} & 4942/ 706/ 1412  & 15.6 & 120,119\\
    Luo (\texttt{luo}) & Nilo-Saharan &East& 4M  & \mafand~\cite{adelani_mafand} & 5161/ 737/ 1474  & 11.7 & 229,927\\
    Mossi (\texttt{mos}) & NC / Gur &West& 8M & \mafand~\cite{adelani_mafand} & 4532/ 648/ 1294  & 9.2 & 168,141\\
    Naija (\texttt{pcm}) & English-Creole &West& 75M & \mafand~\cite{adelani_mafand} & 5646/ 806/ 1613  & 9.4 & 206,404\\
    Chichewa (\texttt{nya}) & NC / Bantu &South-East& 14M & Nation Online Malawi & 6250/ 893/ 1785  & 9.3 & 263,622\\
    \shona (\texttt{sna}) & NC / Bantu &South& 12M & VOA Shona & 6207/ 887/ 1773  & 16.2 & 195,834\\
    Kiswahili (\texttt{swa}) & NC / Bantu &East \& Central & 98M & VOA Swahili & 6593/ 942/ 1883 & 12.7 & 251,678\\
    Setswana (\texttt{tsn}) & NC / Bantu &South& 14M & \mafand~\cite{adelani_mafand} & 3489/ 499/ 996 & 8.8 & 141,069\\
    Akan/Twi (\texttt{twi}) & NC / Kwa &West& 9M & \mafand~\cite{adelani_mafand} & 4240/ 605/ 1211  & 6.3 & 155,985\\
    Wolof (\texttt{wol}) & NC / Senegambia &West& 5M & \mafand~\cite{adelani_mafand} & 4593/ 656/ 1312  & 7.4 & 181,048\\
    \xhosa (\texttt{xho}) & NC / Bantu &South& 9M & Isolezwe Newspaper & 5718/ 817/ 1633  & 15.1 & 127,222\\
    \yoruba (\texttt{yor}) & NC / Volta-Niger &West& 42M & Voice of Nigeria and Asejere & 6877/ 983/ 1964  & 11.4 & 244,144\\
    \zulu (\texttt{zul}) & NC / Bantu &South& 27M & Isolezwe Newspaper & 5848/ 836/ 1670 & 11.0 & 128,658\\
    \bottomrule
  \end{tabular}
  }
  \caption{\textbf{Languages and Data Splits for MasakhaNER 2.0 Corpus}. Language, family (NC: Niger-Congo), number of speakers, news source, and data split in number of sentences}
  \vspace{-4mm}
  \label{tab:languages}
  \end{center}
\end{table*}

\subsection{Focus Languages}
\autoref{tab:languages} provides an overview of the languages in our MasakhaNER 2.0 corpus. We focus on 20 Sub-Saharan African languages\footnote{Our selection was also constrained by the availability of volunteers that speak the languages in different NLP/AI communities in Africa.} with varying numbers of speakers (between 1M--100M) that are spoken by over 500M people in around 27 countries in the Western, Eastern, Central and Southern regions of Africa. The selected languages cover four language families. 17 languages belong to the Niger-Congo language family, and one language belongs to each of the Afro-Asiatic (Hausa), Nilo-Saharan (Luo), and English Creole (Naija) families. Although many languages belong to the Niger-Congo language family, they have different linguistic characteristics. For instance, Bantu languages (eight in our selection) make extensive use of affixes, unlike many languages of non-Bantu subgroups such as Gur, Kwa, and Volta-Niger.

\subsection{Language Characteristics}
\paragraph{Script and Word Order} African languages mainly employ four major writing scripts: Latin, Arabic, N'ko and Ge'ez. 
Our focus languages mostly make use of the Latin script. 
While N'ko is still actively used by the Mande languages like Bambara, the most widely used writing script for the language is Latin. However, some languages use additional letters that go beyond the standard Latin script, e.g., ``\textepsilon'', ``\textopeno'', ``\textipa{\ng}'', ``{\d e}'', and more than one character letters like ``bv'', ``gb'', ``mpf'', ``ntsh''. 17 of the languages are tonal except for Naija, \swahili and Wolof.
Nine of the languages make use of diacritics (e.g., {\'e}, \"{e}, \~{n}). All languages use the SVO word order, while Bambara additionally uses the SOV word order. 

\paragraph{Morphology and Noun classes} Many African languages are morphologically rich. According to the World Atlas of Language Structures~\cite[WALS;][]{wals-59}, 16 of our languages employ strong prefixing or suffixing inflections. Niger-Congo languages are known for their system of noun classification. 12 of the languages \textit{actively} make use of between 6--20 noun classes, including all Bantu languages, \ghomala, Mossi, Akan and Wolof~\cite{Van_de_Velde2006-fz,Payne2017,Bodomo2002TheMO,BabouLoporcaro+2016+1+57}. While noun classes are often marked using affixes on the head word in Bantu languages, some non-Bantu languages, e.g., Wolof make use of a dependent such as a determiner that is not attached to the head word. 
For the other Niger-Congo languages such as Fon, Ewe, Igbo and \yoruba, the use of noun classes is merely \textit{vestigial}~\cite{Konoshenko2019AMS}. 
Three of our languages from the Southern Bantu family (\shona, \xhosa and \zulu) capitalize proper names after the noun class prefix as in the language names themselves. This characteristic may limit transfer from languages without this feature as NER models overfit on capitalization~\cite{mayhew-etal-2019-ner}. 
\autoref{sec:lang_charateristics} provides more details regarding the languages' linguistic characteristics.



\section{MasakhaNER 2.0 Corpus}

\subsection{Data source and collection}

We annotate news articles from local sources. The choice of the news domain is based on the availability of data for many African languages and the variety of named entities types (e.g., person names and locations) as illustrated by popular datasets such as CoNLL-03~\cite{tjong-kim-sang-de-meulder-2003-introduction}.\footnote{We also considered using Wikipedia as our data source, but did not due to quality issues~\cite{alabi-etal-2020-massive}.}
\autoref{tab:languages} shows the sources and sizes of the data we use for annotation. Overall, we collected between 4.8K--11K sentences per language from either a monolingual or a translation corpus. 

\paragraph{Monolingual corpus} We collect a large monolingual corpus for nine languages, mostly from local news articles except for \shona and \swahili texts, which were crawled from Voice of America (VOA) websites.\footnote{\url{www.voashona.com/} and \url{www.voaswahili.com/}}
As \yoruba text was missing diacritics, we asked native speakers to manually add diacritics before annotation. During data collection, we ensured that the articles are from a variety of topics e.g. politics, sports, culture, technology, society, and education. In total, we collected between 8K--11K sentences per language.

\paragraph{Translation corpus} For the remaining languages for which we were unable to obtain sufficient amounts of monolingual data, we use a translation corpus, MAFAND-MT~\cite{adelani_mafand}, which consists of French and English news articles translated into 11 languages. We note that translationese may lead to undesired properties, e.g., unnaturalness. However, we did not observe serious issues during the annotation. The number of sentences is constrained by the size of the MAFAND-MT corpus, which is between 4,800--8,000. 


\subsection{NER Annotation Methodology}
\label{sec:ner_annotation}
We annotated the collected monolingual texts with the ELISA annotation tool~\cite{lin-etal-2018-platforms} with four entity types: Personal name (\texttt{PER}), Location (\texttt{LOC}), Organization (\texttt{ORG}), and date and time (\texttt{DATE}), similar to MasakhaNER 1.0~\cite{adelani-etal-2021-masakhaner}. We made use of the MUC-6 annotation guide.\footnote{\url{https://cs.nyu.edu/~grishman/muc6.html}} 
The annotation was carried out by three native speakers per language recruited from AI/NLP communities in Africa. To ensure high-quality annotation, we recruited a language coordinator to supervise  annotation in each language. We organized two online workshops to train language coordinators on the NER annotation. As part of the training, each coordinator annotated 100 English sentences, which were verified. Each coordinator then trained three annotators in their team using both English and African language texts with the support of the workshop organizers. All annotators and language coordinators received appropriate remuneration.\footnote{\$10 per hour, annotating about 200 sentences per hour.}

At the end of annotation, language coordinators worked with their team to resolve disagreements using the adjudication function of ELISA, which ensures a high inter-annotator agreement score.

\subsection{Quality Control}
\label{lab:quality_control}

\begin{table}[tb]
\small
\centering
 \setlength\tabcolsep{4pt}
\begin{tabular}{lrc||lrc}
\toprule
& \textbf{Fleiss'} & \textbf{QC flags} & & \textbf{Fleiss'} & \textbf{QC flags} \\
\textbf{Lang.} & \textbf{Kappa} & \textbf{fixed?} & \textbf{Lang.} & \textbf{Kappa} & \textbf{fixed?}\\
\midrule
bam               & 0.980 &  \xmark  & pcm & 0.966 & \xmark\\
bbj               & 1.000 &  \cmark  & nya & 0.988 &\cmark \\
ewe               & 0.991 &  \cmark  & sna & 0.957 & \cmark\\
fon               & 0.941 &  \xmark  & swa & 0.974 & \cmark\\
hau               & 0.950 & \xmark   & tsn & 0.962 & \xmark\\
ibo               & 0.965 & \xmark   & twi & 0.932 & \xmark\\
kin               & 0.943 & \xmark   & wol & 0.979 & \cmark\\
lug               & 0.950 & \cmark   & xho & 0.945 & \cmark\\
luo               & 0.907 & \xmark   & yor  & 0.950 &\cmark \\
mos               & 0.927 & \xmark   & zul  & 0.953 & \cmark \\
\bottomrule

\end{tabular}
\caption{Inter-annotator agreement for our datasets calculated using Fleiss' kappa $\kappa$ at the entity level before adjudication. QC flags (\cmark) are the languages that fixed the annotations for all \textbf{Q}uality \textbf{C}ontrol flagged tokens.}  
\label{tab:agreement}
\end{table}


As discussed in \autoref{sec:ner_annotation}, language coordinators helped resolve several disagreements in annotation prior to quality control.  \autoref{tab:agreement} reports the Fleiss Kappa score after the intervention of language coordinators (i.e. post-intervention score). The pre-intervention Fleiss Kappa score was much lower. For example, for \texttt{pcm}, the pre-intervention Fleiss Kappa score was 0.648 and improved to 0.966 after the language coordinator discussed the disagreements with the annotators. 

For the quality control, annotations were automatically adjudicated when there was agreement, but were flagged for further review when annotators disagreed on mention spans or types. 
The process for reviewing and fixing quality control issues was voluntary and so not all languages were further reviewed (see Table \ref{tab:agreement}).

 
We automatically identified positions in the annotation that were more likely to be annotation errors and flagged them for further review and correction. 
The automatic process flags tokens that are commonly annotated as a named entity 
but were not marked as a named entity in a specific position. 
For example, the token \emph{Province} may appear commonly as part of a named entity and infrequently not as a named entity, so when it is seen as not marked it was flagged. Similarly, we flagged tokens that had near-zero entropy with regard to a certain entity type, for example a token almost always annotated as ORG but very rarely annotated as PER.
We also flagged potential sentence boundary errors by identifying sentences with few tokens or sentences which end in a token that appears to be an abbreviation or acronym. 
As shown in Table \ref{tab:agreement}, before further adjudication and correction there was already relatively high inter-annotator agreement measured by Fleiss' Kappa at the mention level.

After quality control, we divided the annotation into training, development, and test splits consisting of 70\%, 10\%, and 20\% of the data respectively. \autoref{sec:appendix_data_source} provide details on the number of tokens per entity (PER, LOC, ORG, and DATE) and the fraction of entities in the tokens. 

\begin{table}[t]
 \begin{center}
 \setlength\tabcolsep{3pt}
 \scalebox{0.82}{
 \footnotesize
  \begin{tabular}{p{29mm}rp{48mm}}
    \toprule
    \textbf{PLM} & \textbf{\# Lang.} & \textbf{Languages in MasakhaNER 2.0} \\
    \midrule
    mBERT-cased (110M) &104 & \textbf{swa}, \textbf{yor}  \\
    \addlinespace[0.5em]
    
    XLM-R-base/large (270M / 550M) &100 & \textbf{hau}, \textbf{swa}, \textbf{xho}  \\
    \addlinespace[0.5em]
    
    mDeBERTaV3 (276M) & 100 & \textbf{hau}, \textbf{swa}, \textbf{xho}  \\
    \addlinespace[0.5em]
    
    RemBERT (575M) &110  & \textbf{hau}, \textbf{ibo}, \textbf{nya}, \textbf{sna}, \textbf{swa}, \textbf{xho}, \textbf{yor}, \textbf{zul}  \\
    
    \addlinespace[0.5em]
    
    AfriBERTa (126M) & 11 &  \textbf{hau}, \textbf{ibo}, \textbf{kin}, \textbf{pcm}, \textbf{swa}, \textbf{yor}  \\
    
    \addlinespace[0.5em]
    AfroXLMR-base/large (270M/550M) & 20 & \textbf{hau}, \textbf{ibo}, \textbf{kin}, \textbf{nya}, \textbf{pcm}, \textbf{sna},  \textbf{swa}, \textbf{xho}, \textbf{yor}, \textbf{zul}   \\

    \bottomrule
  \end{tabular}
  }
  \vspace{-3mm}
  \caption{Language coverage and size for PLMs.}
  \label{tab:plm_languages}
  \vspace{-3mm}
  \end{center}
\end{table}

\section{Baseline Experiments}

\subsection{Baseline Models}
As baselines, we fine-tune several multilingual PLMs including mBERT~\cite{bert}, XLM-R~\cite[base \& large;][]{conneau-etal-2020-unsupervised}, mDeBERTaV3~\cite{He2021DeBERTaV3ID}, AfriBERTa~\cite{ogueji-etal-2021-small}, RemBERT~\cite{chung2021rethinking}, and AfroXLM-R~\cite[base \& large;][]{alabi-etal-2022-adapting}. We fine-tune the PLMs on each language's training data and evaluate performance on the test set using HuggingFace Transformers 
\cite{wolf-etal-2020-transformers}.

\paragraph{Massively multilingual PLMs}
\autoref{tab:plm_languages} shows the language coverage and size of different massively multilingual PLMs trained on 100--110 languages. mBERT was pre-trained using masked language modeling (MLM) and next-sentence prediction on 104 languages, including \texttt{swa} and \texttt{yor}. RemBERT was trained with a similar objective, but makes use of a larger output embedding size during pre-training and covers more African languages. XLM-R was trained only with MLM on 100 languages and on a larger pre-training corpus. mDeBERTaV3 makes use of ELECTRA-style~\cite{clark2020electra} pre-training, i.e., a replaced token detection (RTD) objective instead of MLM. 

\paragraph{Africa-centric multilingual PLMs}
We also obtained NER models by fine-tuning two PLMs that are pre-trained on African languages.
AfriBERTa~\citep{ogueji-etal-2021-small} was pre-trained on less than 1 GB of text covering 11 African languages, including six of our focus languages, and has shown impressive performance on NER and sentiment classification for languages in its pre-training data~\cite{adelani-etal-2021-masakhaner, Muhammad2022NaijaSentiAN}. AfroXLM-R~\cite{alabi-etal-2022-adapting} is a language-adapted \cite{pfeiffer-etal-2020-mad} version of XLM-R that was fine-tuned on 17 African languages and three high-resource languages widely spoken in Africa (``eng'', ``fra'', and ``ara''). \autoref{sec:model_reproducibility} provides the model hyper-parameters for fine-tuning the PLMs. 

\subsection{Baseline Results}

\begin{table*}[t]
\begin{center}
\footnotesize
\resizebox{\textwidth}{!}{%
\begin{tabular}{lrrrrrrrrrrrrrrrrrrrr|c}
\toprule
\textbf{Model} & \textbf{bam} & \textbf{bbj} & \textbf{ewe} & \textbf{fon} & \textbf{hau} & \textbf{ibo} & \textbf{kin} & \textbf{lug} & \textbf{luo} & \textbf{mos} & \textbf{nya} & \textbf{pcm} & \textbf{sna} & \textbf{swa} & \textbf{tsn} & \textbf{twi}  & \textbf{wol} & \textbf{xho} & \textbf{yor} & \textbf{zul} & \textbf{AVG}  \\
\midrule
\multicolumn{7}{l}{\textit{PLM pre-trained on 100+ world languages}} \\
mBERT & 78.9 & 60.6 & 86.9 & 79.9 & 85.2 & 87.3 & 83.2 & 85.5 & 80.3 & 71.4 & 88.6 & 87.1 & 92.4 & 92.1 & 86.4 & 75.7 & 79.9 & 85.0 & 87.7 & 81.7 & $82.8_{\pm0.2}$ \\ 
XLM-R-base & 78.7 & 72.3 & 88.5 & 81.9 & 83.8 & 87.8 & 82.5 & 86.7 & 79.3 & 72.7 & 89.9 & 88.5 & 93.6 & 92.2 & 86.1 & 78.7 & 82.3 & 87.0 & 85.8 & 84.6 & $84.1_{\pm0.1}$ \\ 
XLM-R-large & 79.4 & \textbf{75.2} & 89.1 & 81.6 & 86.3 & 87.2 & 84.3 & 88.1 & 80.8 & 74.9 & 90.5 & 89.2 & 94.2 & 92.6 & 85.9 & 79.8 & 82.0 & 88.1 & 86.6 & 86.7 & $85.1_{\pm0.5}$ \\ 
RemBERT & 80.1 & 74.2 & 89.2 & 82.2 & 84.7 & 86.4 & 85.2 & 87.1 & 80.4 & 72.7 & 91.4 & 89.5 & 94.8 & 92.0 & 87.0 & 78.5 & 83.6 & 88.3 & 87.2 & 85.5 & $85.0_{\pm0.2}$ \\ 
mDeBERTaV3 & 80.2 & 73.5 & 89.8 & 81.8 & 85.4 & 88.8 & 86.4 & 88.7 & 80.3 & \textbf{76.4} & 92.0 & \textbf{90.1} & 95.5 & 92.5 & 86.5 & 79.4 & 83.6 & 88.1 & 86.7 & 88.3 & $85.7_{\pm0.2}$ \\ 
\midrule
\multicolumn{7}{l}{\textit{PLM pre-trained on African languages}} \\
AfriBERTa & 78.6 & 71.0 & 86.9 & 79.9 & 85.2 & 87.3 & 83.2 & 85.5 & 78.4 & 71.4 & 88.6 & 87.1 & 92.4 & 92.1 & 83.2 & 75.7 & 79.9 & 85.0 & 87.7 & 81.7 & $83.0_{\pm0.2}$ \\ 
AfroXLMR-base & 79.6 & 73.3 & 89.2 & 82.3 & 86.6 & 88.5 & 86.1 & 88.1 & 80.8 & 74.4 & 91.9 & 89.3 & 95.7 & 92.3 & 87.7 & 78.9 & 84.9 & 88.6 & 88.3 & 88.4 & $85.7_{\pm0.1}$ \\ 
AfroXLMR-large & \textbf{82.2} & 74.8 & \textbf{90.3} & \textbf{82.7} & \textbf{87.4} & \textbf{89.6} & \textbf{87.5} & \textbf{89.6} & \textbf{82.2} & \textbf{76.4} & \textbf{92.4} & 89.7 & \textbf{96.2} &\textbf{92.7} & \textbf{89.4} & \textbf{81.1} & \textbf{86.8} & \textbf{89.9} & \textbf{89.3} & \textbf{90.6} & $\mathbf{87.0_{\pm0.2}}$ \\ 
\bottomrule
    \end{tabular}
    }
\caption{\textbf{NER Baselines on MasakhaNER 2.0}. We compare several multilingual PLMs including the ones trained on African languages. Average is over 5 runs. }
\label{tab:masakhaner_baselines}
  \end{center}
\end{table*}

\begin{table*}[!ht]
\begin{center}
\footnotesize
\resizebox{\textwidth}{!}{%
\begin{tabular}{llrrrrrrrrrrrrrrrrrrrr|c}
\toprule
\textbf{Train Lang.} & \textbf{Data}& \textbf{bam} & \textbf{bbj} & \textbf{ewe} & \textbf{fon} & \textbf{hau} & \textbf{ibo} & \textbf{kin} & \textbf{lug} & \textbf{luo} & \textbf{mos} & \textbf{nya} & \textbf{pcm} & \textbf{sna} & \textbf{swa} & \textbf{tsn} & \textbf{twi}  & \textbf{wol} & \textbf{xho} & \textbf{yor} & \textbf{zul} & \textbf{AVG}  \\
\midrule
\multicolumn{2}{c}{Language in MasakhaNER 1.0?} & \xmark & \xmark & \xmark & \xmark & \cmark & \cmark & \cmark & \cmark & \cmark & \xmark & \xmark & \cmark & \xmark & \cmark & \xmark & \xmark & \cmark & \xmark & \cmark & \xmark & - \\ 
\midrule
\multicolumn{7}{l}{\textit{Evaluation on MasakhaNER 2.0 test set}} \\
(a) MasakhaNER 1.0 & MasakhaNER 1.0 & 52.2 & 48.4 & 78.3 & 52.9 & 76.9 & 86.0 & 77.6 & 83.2 & 68.6 & 55.0 & 82.1 & 86.7 & 49.6 & 89.4 & 80.0 & 56.6 & 73.6 & 56.9 & 69.4 & 69.9 & $69.7_{\pm 0.6}$ \\

(b) MasakhaNER 1.0 & MasakhaNER 2.0 & 50.9 & 49.8 & 76.2 & 57.1 & 88.7 & 90.1 & 87.6 & 90.0 & 82.7 & 49.6 & 80.4 & 90.2 & 42.5 & 93.1 & 79.4 & 57.3 & 87.0 & 47.4 & 89.7 & 64.3 & $72.7_{\pm 0.6}$ \\

(c) MasakhaNER 2.0 & MasakhaNER 2.0 & \textbf{82.3} & \textbf{75.5} & \textbf{89.5} & \textbf{83.2} & \textbf{87.7} & \textbf{92.3} & \textbf{87.2} & \textbf{89.1} & \textbf{81.8} & \textbf{75.3} & \textbf{92.2} & \textbf{89.9} & \textbf{95.9} & \textbf{93.1} & \textbf{89.5} & \textbf{78.8} & \textbf{86.4} & \textbf{89.7} & \textbf{89.1} & \textbf{90.7} & $\mathbf{87.0_{\pm 1.2}}$ \\

\midrule
\multicolumn{7}{l}{\textit{Evaluation on MasakhaNER 1.0 test set}} \\
(a) MasakhaNER 1.0 & MasakhaNER 1.0 & -- & -- & -- & -- & 92.1 & 89.2 & 79.1 & 86.0 & 80.0 & -- & -- & 91.2 & -- & 89.5 & -- & -- & 70.8 & -- & 85.0 & -- & $84.8_{\pm 0.3}$ \\

(b) MasakhaNER 1.0 & MasakhaNER 2.0 & -- & -- & -- & -- & 80.8 & 84.6 & 77.7 & 79.0 & 67.0 & -- & -- & 88.0 & -- & 86.3 & -- & -- & 71.6 & -- & 85.0 & -- & $80.0_{\pm 0.3}$ \\
(c) MasakhaNER 2.0 & MasakhaNER 2.0 & -- & -- & -- & -- & 80.4 & 84.3 & 77.0 & 79.8 & 67.6 & -- & -- & 87.9 & -- & 86.5 & -- & -- & 72.1 & -- & 84.8 & -- & $80.1_{\pm 0.8}$ \\
\bottomrule
\end{tabular}
}
\caption{\textbf{Multilingual evaluation on African NER datasets}. We compare the performance of AfroXLM-R-large trained on languages of MasakhaNER 2.0 and MasakhaNER 1.0 and evaluated both on the same and on the other dataset. The first column indicate the languages used for training (the 10 languages from MasakhaNER or the 20 languages from MasakhaNER 2.0). The second column indicates the training data. Average is over 5 runs. 
}
\label{tab:masakhaner_comparison_10v20}
\end{center}
\end{table*}

\autoref{tab:masakhaner_baselines} shows the results of training NER models on each language using the eight multilingual and Africa-centric PLMs. All PLMs provided good performance in general. However, we observed worse results for mBERT and AfriBERTa especially for languages they were not pre-trained on. For instance, both models performed between 6--12 F1 worse for \texttt{bbj}, \texttt{wol} or \texttt{zul} compared to XLM-R-base. We hypothesize that the performance drop is largely due to the small number of African languages covered by mBERT as well as AfriBERTa's comparatively small model capacity. XLM-R-base gave much better performance ($>1.0$ F1) on average compared to mBERT and AfriBERTa. We found the larger variants of mBERT and XLM-R, i.e., RemBERT and XLM-R-large to give much better performance ($>2.0$ F1) than the smaller models. Their larger capacity facilitates positive transfer, yielding better performance for unseen languages. Surprisingly, mDeBERTaV3 provided slightly better results than XLM-R-large and RemBERT despite its smaller size, demonstrating the benefits of the RTD pre-training~\cite{clark2020electra}. 

The best PLM is AfroXLM-R-large, which outperforms mDeBERTaV3, RemBERT and AfriBERTa by $+1.3$ F1, $+2.0$ F1 and $+4.0$ F1 respectively. Even the performance of its smaller variant, AfroXLM-R-base is comparable to mDeBERTaV3. Overall, our baseline results highlight that large PLMs, PLM with improved pre-training objectives, and PLMs pre-trained on the target African languages are able to achieve reasonable baseline performance. Combining these criteria provides improved performance, such as AfroXLM-R-large, a large PLM trained on several African languages. 

\subsection{Entity-level Analysis of MasakhaNER 2.0}

\subsubsection{Error Analysis with ExplainaBoard}
Furthermore, using ExplainaBoard~\cite{liu-etal-2021-explainaboard}, we analysed the best three baseline NER models: AfroXLM-R-large, mDeBERTaV3, and XLM-R-large. 
We discovered that 2-token entities were easier to predict accurately than lengthier entities (4 or more words). Moreover, the result shows that all the models have difficulty predicting 
zero-frequency entities effectively (entities with no occurrences in the training set). 
Interestingly, AfroXLMR-large is significantly better than other models for zero-frequency entities, suggesting that training PLMs on African languages promotes generalization to unseen entities. Finally, we observed that the three models perform better when predicting PER and LOC entities compared to ORG and DATE entities by up to (+5\%).  
\autoref{sec:error_analysis_app} provides more details on the error analysis. 


\subsubsection{Dataset Geography of Entities}
\begin{figure*}[t]
\centering
    \hspace{-2.5\baselineskip}
    \begin{subfigure}{0.42\textwidth}
       \centering
    	\includegraphics[width=0.95\linewidth]{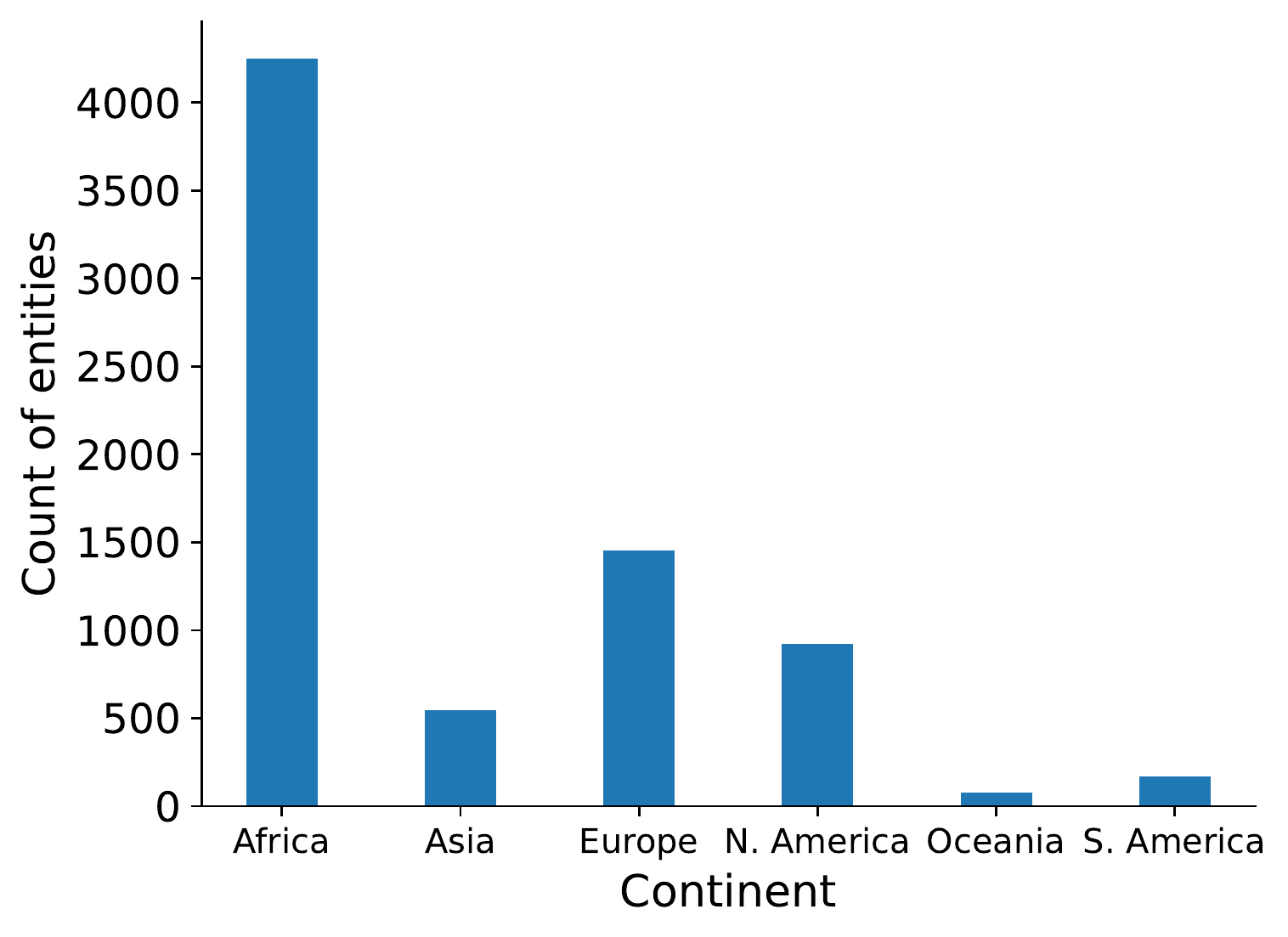}
        \caption{Number of entities per continent} \label{fig:continents}
    \end{subfigure}
   \begin{subfigure}{0.42\textwidth}
    \centering
	\includegraphics[width=0.85\linewidth]{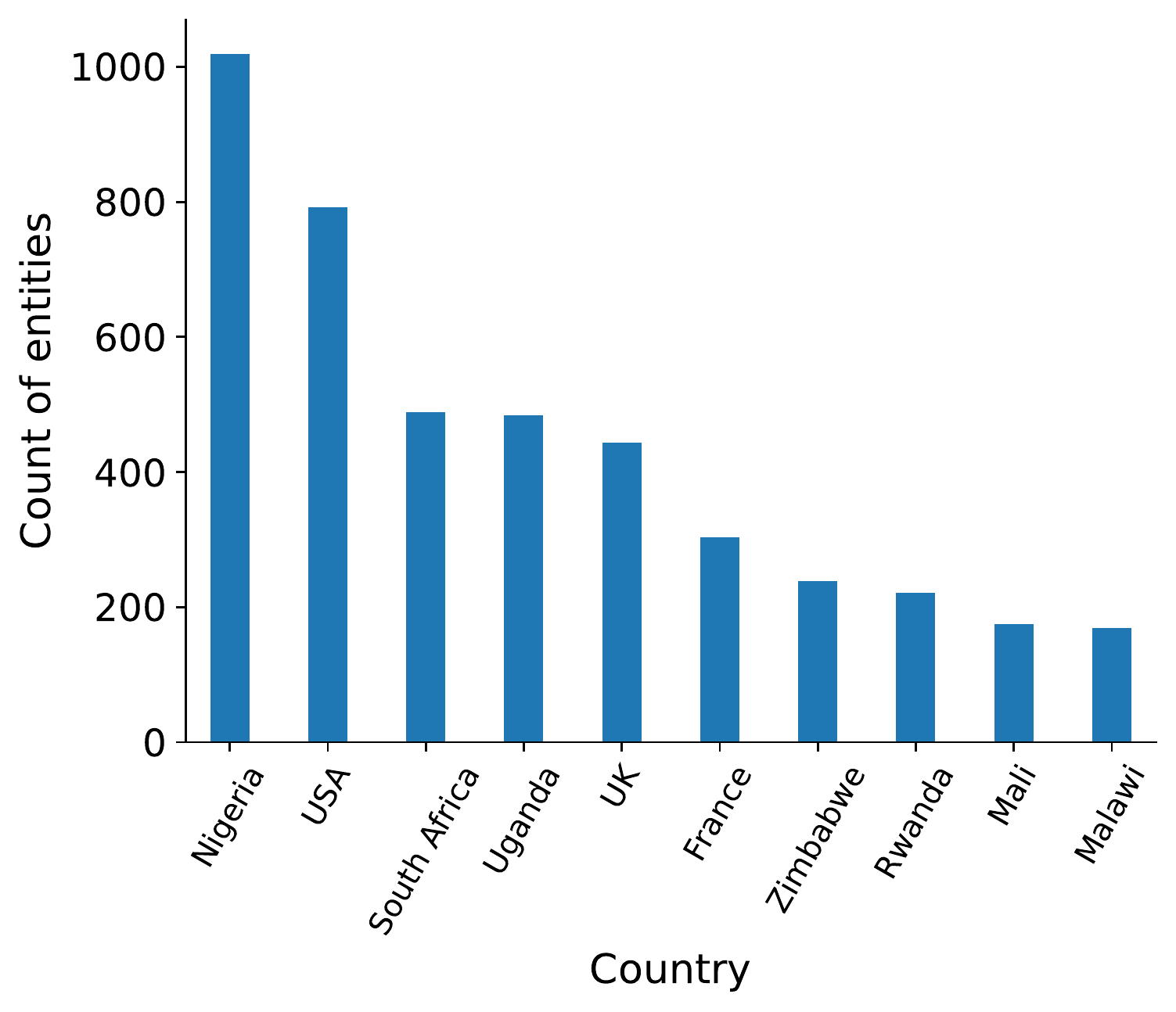}
    \caption{Top-10 countries} \label{fig:top_10}
    \end{subfigure}
    \hspace{-3\baselineskip}
    \vspace{-3mm}
    \caption{\textbf{Number of entities per continent and the top-10 countries with the largest number of entities}}
    \label{fig:data_geography}
    \vspace{-3mm}
\end{figure*}

Next, we analyse the geographical representativeness of the entities in our dataset, specifically, we measure the count of entities based on the countries they originate from. Following the approach of \citet{faisal-etal-2022-dataset}, we first performed entity linking of named entities present in our dataset to Wikidata IDs using mGenre~\cite{de-cao-etal-2022-multilingual}, followed by mapping Wikidata IDs to countries.  

\autoref{fig:data_geography} shows the result of number of entities per continent and the top-10 countries with the largest representation of entities. Over 50\% of the entities are from Africa, followed by Europe. This shows that the entities of MasakhaNER 2.0 properly represent the African continent. Seven out of the top-10 countries are from Africa, but also includes USA, United Kingdom and France. 

\subsection{Transfer Between African NER Datasets}

African languages have a diverse set of linguistic characteristics.  
To demonstrate this heterogeneity, we perform a transfer learning experiment where we compare the performance of multilingual NER models jointly trained on the languages of MasakhaNER 1.0 or MasakhaNER 2.0 and perform zero-shot evaluation on both test sets. We consider three experimental settings:
\begin{enumerate}[label=(\alph*)]
    \item Train on all languages in MasakhaNER 1.0 using MasakhaNER 1.0 training data. 
    \item Train on the languages in MasakhaNER 1.0 (excl. ``amh'') using the MasakhaNER 2.0 training data.
    \item Train on all languages in MasakhaNER 2.0 using MasakhaNER 2.0 training data.
\end{enumerate}
\autoref{tab:masakhaner_comparison_10v20} shows the result of the three settings. When evaluating on the MasakhaNER 2.0 test set in setting (a), the performance is mostly high ($>65$ F1) for languages in MasakhaNER 1.0. Most of the languages that are not in MasakhaNER 1.0 have worse zero-shot performance, typically between $48-60$ F1 except for \texttt{ewe}, \texttt{nya}, \texttt{tsn}, and \texttt{zul} with over $69$ F1. Making use of a larger dataset, i.e., setting (b) from MasakhaNER 2.0 only provides a small improvement ($+3$ F1). The evaluation on setting (c) shows a large gap of about $15$ F1 and $17$ F1 compared to settings (b) and (a) on the MasakhaNER 2.0 test set respectively, especially for Southern Bantu languages like \texttt{sna} and \texttt{xho}. On the MasakhaNER 1.0 test set, training on the in-distribution MasakhaNER 1.0 languages and training set achieves the best performance. However, the performance gap compared to training on the MasakhaNER 2.0 data is much smaller. 
Overall, these results demonstrate the need to create large benchmark datasets (like MasakhaNER 2.0) covering diverse languages with different linguistic characteristics, particularly for the Africa.

\begin{figure}[t]
    \centering
    \includegraphics[width=1.02\linewidth]{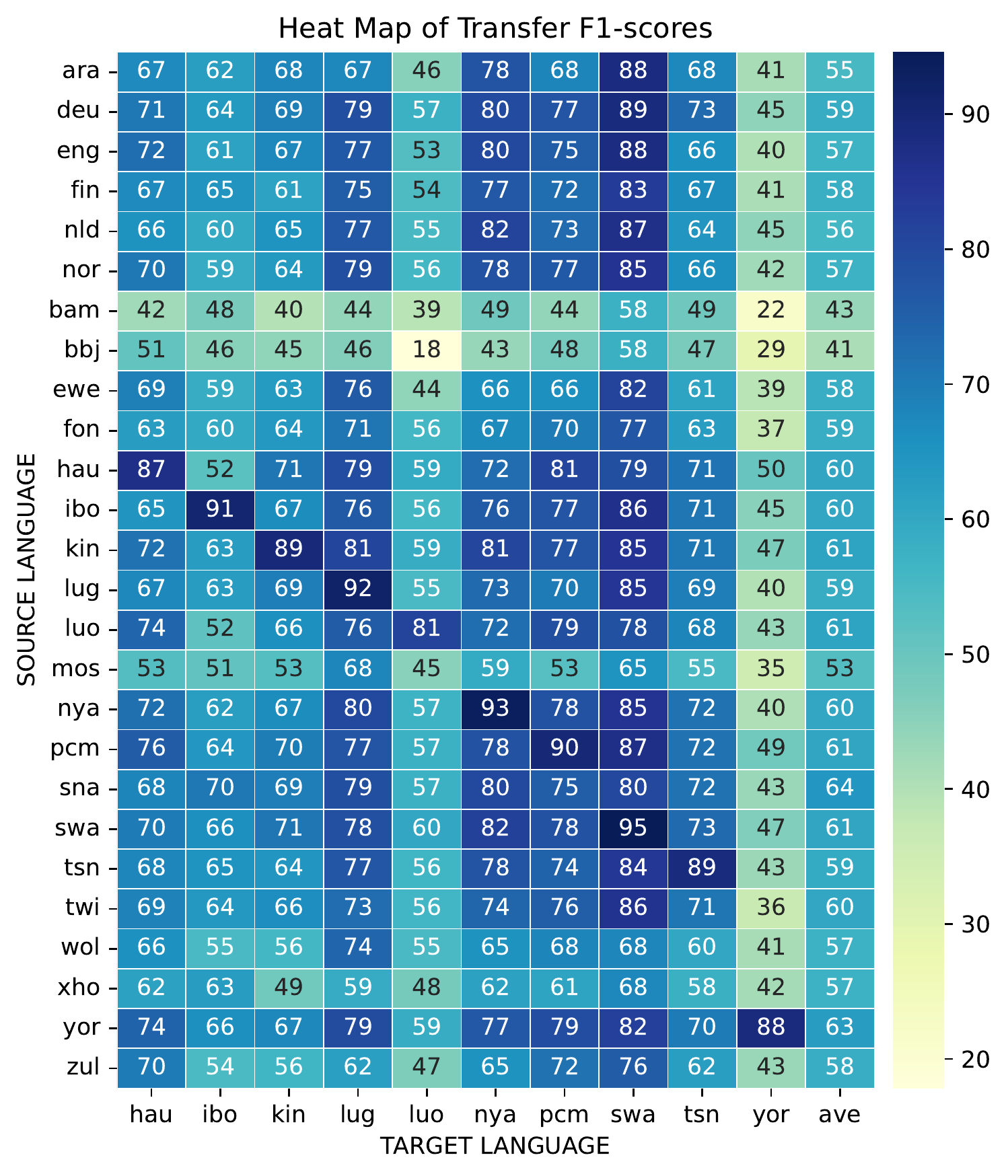}
    \vspace{-7mm}
    \caption{\textbf{Zero-shot Transfer} from several source languages to African languages for 10 languages in MasakhaNER 2.0 and the average (ave) over all 20 languages. Appendix \ref{sec:zero_transfer_other_lang} shows results for each of the 20 languages.}
    \label{fig:transfer_image_0_shot}
    \vspace{-4mm}
\end{figure}

\section{Cross-Lingual Transfer}

\begin{table*}[t]
 \begin{center}
\resizebox{\textwidth}{!}{%
 \footnotesize
  \begin{tabular}{llllrrrrrrr}
    \toprule
        
    \textbf{} & \textbf{Top-2} & \textbf{Top-2} & \textbf{Top-3 features selected} & \textbf{Target} & \textbf{Top-1} & \textbf{Top-2} & \textbf{Top-2} & \textbf{Best} & \textbf{Second} & \textbf{eng} \\
    
    \textbf{Target} & \textbf{Transf.} & \textbf{LangRank} & \textbf{by LangRank model} & \textbf{Lang.} & \textbf{LangRank} & \textbf{LangRank} & \textbf{Transf.} & \textbf{Transf.} & \textbf{Best} & \textbf{Tranf.} \\
    
    \textbf{Lang.} & \textbf{Lang} & \textbf{Model} & \textbf{Lang 1; Lang 2} & \textbf{F1} & \textbf{Lang. F1} & \textbf{Lang. F1} & \textbf{Lang. F1} & \textbf{F1} & \textbf{Transf. F1} & \textbf{F1}  \\
    \midrule
    
    \textbf{bam} & twi, fon & wol, fon & $(d_{geo}, d_{inv}, sr); (d_{geo}, sr, d_{pho})$ & 80.4 & 47.1 & 52.8 & \textbf{\underline{55.1}} & 54.3 & 53.0 & 38.4 \\ 
    
    \textbf{bbj} & fon, ewe & twi, ewe & $(s_{tf}, d_{syn}, d_{geo}); (s_{tf}, d_{geo}, sr)$ & 72.9 & 53.9 & 58.8 & \textbf{\underline{60.1}} & 59.8 & 58.4 & 45.8 \\
    
    \rowcolor{Color}
    \textbf{ewe} & swa, twi & pcm, swa & $(d_{geo}, s_{tf}, sr); (eo, d_{geo}, s_{tf})$ & 91.7 & 78.1 & 81.1 & \textbf{\underline{83.9}} & 81.6 & 81.5 & 76.4 \\ 
    
    \textbf{fon} & mos, bbj & yor, ewe & $(d_{geo}, d_{syn}, sr); (s_{tf}, d_{geo}, d_{gen})$ & 84.9 & 58.4 & 64.9 & \textbf{\underline{69.9}} & 65.4 & 62.0 & 50.6 \\
    
    \rowcolor{Color}
    \textbf{hau} & pcm, yor & yor, swa & $(d_{geo}, sr, eo); (eo, sr, s_{tf})$ & 86.9 & 74.3 & 74.8 & \textbf{\underline{77.4}} & 75.9 & 74.3 & 72.4  \\ 
    
    \rowcolor{Color}
    \textbf{ibo} & sna, yor & pcm, kin & $(eo, d_{geo}, s_{tf}); (d_{geo}, sr, eo)$ & 91.0 & 64.2 &  63.9 & \textbf{\underline{77.1}} & 70.4 & 66.0 & 61.4 \\ 
    
    \rowcolor{Color}
    \textbf{kin} & hau, swa & sna, yor & $(eo, d_{geo}, s_{tf}); (eo, s_{tf}, sr)$ & 89.5 & 69.2 & 71.8 & \textbf{\underline{74.0}} & 71.1 & 70.6 & 67.4 \\
    
    \rowcolor{Color}
    \textbf{lug} & kin, nya & luo, zul & $(d_{geo}, sr, eo); (d_{syn}, d_{geo}, sr)$ & 91.5 & 75.9 & 78.1 & \textbf{\underline{82.1}} & 81.1 & 80.0 & 76.5 \\ 
    
    \textbf{luo} & swa, hau & lug, sna & $(d_{geo}, sr, eo); (d_{geo}, eo, sr)$ & 81.2 & 54.9 & 61.6 & \textbf{\underline{61.1}} & 60.4 & 59.5 & 53.4 \\
    
   \textbf{mos} & fon, ewe & yor, fon & $(d_{geo}, d_{inv}, sr); (d_{geo}, s_{tf}, sr)$ & 78.9 & 50.8 & 62.5 & \textbf{\underline{65.6}} & 64.2 & 60.4 & 45.4 \\
   
   \rowcolor{Color}
    \textbf{nya} & swa, nld & zul, sna & $(eo, d_{geo}, sr); (d_{geo}, eo, d_{syn})$ & 93.5 & 65.5 & 81.5 & \textbf{\underline{81.8}} & 81.8 & 81.7 & 80.1 \\
    
    \rowcolor{Color}
    \textbf{pcm} & hau, yor & eng, yor & $(eo, d_{gen}, d_{syn}); (eo, d_{geo}, sr)$ & 89.9 & 75.5 & 79.9 & \textbf{\underline{81.8}} & 80.5 & 79.1 & 75.5 \\ 
    
    \rowcolor{Color}
    \textbf{sna} & zul, xho & swa, zul & $(eo, sr, s_{tf}); (d_{geo}, sr, eo)$ & 96.0 & 32.4 & 80.0 & \textbf{\underline{80.0}} & 77.5 & 74.5 & 37.1 \\
    
    \rowcolor{Color}
    \textbf{swa} & deu, ara & ita, nld & $(sr, d_{inv}, eo); (eo, s_{tf}, sr)$ & 94.6 & 84.5 & 86.0 & \textbf{\underline{89.6}} & 88.7 & 88.1 & 87.9 \\ 
    
    \rowcolor{Color}
    \textbf{tsn} & deu, swa & swa, nya & $(eo, d_{inv}, s_{tf}); (d_{inv}, d_{geo}, d_{gen})$ & 88.7 & 73.1 & 73.4 & \textbf{\underline{74.0}} & 73.3 & 73.1 & 65.8 \\ 
    
    \textbf{twi} & swa, nya & swa, ewe & $(eo, s_{tf}, d_{geo}); (d_{geo}, s_{tf}, sr)$ & 82.0 & 61.9 & 57.2 & \textbf{\underline{64.3}} & 61.0 & 61.9 & 49.5\\
    
    \textbf{wol} & fon, mos & fon, yor & $(d_{geo}, sr, s_{tf}); (sr, d_{geo}, d_{syn})$ & 85.2 & 62.0 &  59.4 & \textbf{\underline{63.0}} & 62.0 & 58.9 & 44.8 \\ 
    
    \rowcolor{Color}
    \textbf{xho} & zul, sna & zul, pcm & $(eo, d_{geo}, d_{gen}); (eo, s_{tf}, d_{inv})$ & 90.8 & 83.7 & 83.0 & \textbf{\underline{84.3}} & 83.7 & 74.0 & 24.5 \\
    
    \textbf{yor} & hau, pcm & fon, pcm & $(d_{geo}, d_{inv}, d_{syn}); (eo, d_{geo}, d_{inv})$ & 88.3 & 37.3 &  43.2 & \textbf{\underline{50.3}} & 50.3 & 48.8 & 40.4 \\
    
    \rowcolor{Color}
    \textbf{zul} & xho, sna & xho, sna & $(eo, d_{gen}, d_{geo}); (d_{syn}, sr, d_{geo})$ & 88.6 & 82.1 & 85.5 & \textbf{\underline{85.5}} & 82.1 & 69.4 & 44.7 \\ 
    \midrule
    AVG & -- & & & 87.3 & 64.2 & 69.8 & \textbf{73.1} & 71.3 & 68.8 & 56.9\\ 

    \bottomrule
  \end{tabular}
  }
  \vspace{-3mm}
  \caption{\textbf{Best Transfer Languages for NER.} The best zero-shot result is \textbf{bolded}, numbers that are not significantly different are \underline{underlined}. The ranking model features are based on the definitions in \cite{lin-etal-2019-choosing} like: geographic distance ($d_{geo}$), genetic distance ($d_{gen}$), inventory distance ($d_{inv}$), syntactic distance ($d_{syn}$), phonological distance ($d_{pho}$), transfer language dataset size ($s_{tf}$), transfer over target size ratio ($sr$), and entity overlap ($eo$). The languages highlighted in gray have very good transfer performance ($>70\%$) using the best transfer language.  
  }
  \label{tab:best_transfer_language}
  \vspace{-3mm}
  \end{center}
\end{table*}
The success of cross-lingual transfer either in zero or few-shot settings depends on several factors, including an appropriate selection of the best source language. Several attempts at cross-lingual transfer make use of English as the source language due to its availability of training data. However, English is unrepresentative of African languages and transfer performance is often lower for distant languages~\citep{adelani-etal-2021-masakhaner}.

\subsection{Choosing Transfer Languages for NER}
Here, we follow the approach of \citet{lin-etal-2019-choosing}, \texttt{LangRank}, that uses source-target transfer evaluation scores and data-dependent features such as dataset size 
and entity overlap, and six different linguistic distance measures based on \texttt{lang2vec}~\cite{littell-etal-2017-uriel} such as geographic distance ($d_{geo}$), genetic distance ($d_{gen}$), inventory distance ($d_{inv}$), syntactic distance ($d_{syn}$), phonological distance ($d_{pho}$), and featural distance ($d_{fea}$).
We provide definitions of the features in \autoref{sec:lang_fea_descr}.  
\texttt{LangRank} is trained using these features to determine the best transfer language in a leave-one-out setting where, for each target language, we train on all other languages except the target language.
We compute transfer F1 scores from a set of $N$ transfer (source) languages and evaluate on $N$ target languages, yielding $N\times N$ transfer scores.

\paragraph{Choice of Transfer Languages} We selected 22 human-annotated NER datasets of diverse languages by searching the web and  HuggingFace Dataset Hub~\cite{lhoest-etal-2021-datasets}. We required each dataset to contain at least the PER, ORG, and LOC types, and we limit our analysis to these types. We also added our MasakhaNER 2.0 dataset with 20 languages. In total, the datasets cover 42 languages (21 African). Each language is associated with a single dataset. \autoref{sec:non_african_ner_data} provides details about the languages, datasets, and data splits. To compute zero-shot transfer scores, we fine-tune mDeBERTaV3 on the NER dataset of a source language and perform zero-shot transfer to the target languages.
We choose mDeBERTaV3 because it supports 100 languages and has the best performance among the PLMs trained on a similar number of languages.

\subsection{Single-source Transfer Results}

\autoref{fig:transfer_image_0_shot} shows the zero-shot evaluation of training on 42 NER datasets and evaluation on the test sets of the 20 MasakhaNER 2.0 languages. On average, we find the transfer from non-African languages to be slightly worse (51.7 F1) than transfer from African languages (57.3 F1). The worst transfer result is using \texttt{bbj} as source language (41.0 F1) while the best is using \texttt{sna} ($64$ F1), followed by \texttt{yor} ($63$ F1).

We identify German (\texttt{deu}) and Finnish (\texttt{fin}) as the top-2 transfer languages among the non-African languages. In most cases, languages that are geographically and syntactically close tend to benefit most from each other. For example, \texttt{sna}, \texttt{xho}, and \texttt{zul} have very good transfer among themselves due to both syntactic and geographical closeness. Similarly, for Nigerian languages (\texttt{hau}, \texttt{ibo}, \texttt{pcm}, \texttt{yor}) and East African languages (\texttt{kin}, \texttt{lug}, \texttt{luo}, \texttt{swa}), geographical proximity plays an important role. While most African languages prefer transfer from another African language, there are few exceptions, like \texttt{swa} preferring transfer from \texttt{deu} or \texttt{ara}. The latter can be explained by the presence of Arabic loanwords in Swahili~\citep{arabic_swahili_influence}. Similarly, \texttt{nya} and \texttt{tsn} also prefer \texttt{deu}. \autoref{sec:zero_transfer_other_lang} provides results for transfer to non-African languages.


\subsection{LangRank and Co-training Results} \label{sec:top2lang}

We also investigate the benefit of training on the second-best language in addition to the languages selected by \texttt{LangRank}. We jointly train on the combined data of the top-2 transfer languages or the top-2 languages predicted by \texttt{LangRank} and evaluate their zero-shot performance on the target language.
\autoref{tab:best_transfer_language} shows the result for the top-2 transfer languages using the best from $42\times 42 $ transfer F1-scores and \texttt{LangRank} model predictions. \texttt{LangRank} predicted the right language as one of the top-2 best transfer language in 13 target languages. The target languages with incorrect predictions are \texttt{fon}, \texttt{ibo}, \texttt{kin}, \texttt{lug}, \texttt{luo}, \texttt{nya}, and \texttt{swa}. The transfer languages predicted as alternative are often in the top-5 transfer languages or are less than ($-5$ F1) worse than the best transfer language. For example, the best transfer language for \texttt{lug} is \texttt{kin} ($81$ F1) but \texttt{LangRank} predicted \texttt{luo} ($76$ F1). \autoref{sec:best_transfer_other_lang} gives results for non-African languages. 

\paragraph{Features that are important for transfer}
The most important features for the selection of best language by \texttt{LangRank} are geographic distance ($d_{geo}$) and entity overlap ($eo$). 
The $d_{geo}$ is influential because named entities (e.g. name of a politician or a city) are often similar from languages spoken in the same country (e.g. Nigeria with 4 languages in MasakhaNER 2.0) or region (e.g. East African languages). Similarly, we find entity overlap to have a positive Spearman correlation ($R=0.6$) to transfer F1-score. \autoref{sec:word_overlap} provides more details on the correlation results. 
$d_{geo}$ occurred as part of the top-3 features for 15 best transfer language and 16 second-best languages. Similarly, for $eo$, it appeared 11--13 times for the top-2 transfer languages. Interestingly, dataset size was not among the most important features, highlighting the need for typologically diverse training data.

\paragraph{Best Transfer Language Outperforms English} We compare the zero-shot transfer performance of the top-2 transfer languages to using \texttt{eng} as the transfer language. They significantly outperform the \texttt{eng} average of $56.9$ by $+14$ and $+12$ F1 for the first and second-best source language, respectively.

\paragraph{Co-training of Top-2 Transfer Languages Improves Performance}
We find that co-training the top-2 transfer languages further improves zero-shot performance over the best transfer by around $+3$ F1. It is most significant for \texttt{fon}, \texttt{ibo}, \texttt{kin} and \texttt{twi} with 3--7 F1 improvement. Co-training the top-2 transfer languages predicted by \texttt{LangRank} is better than using the second-best transfer language, but often performs worse than the best transfer language. 

\begin{figure}[tb]
    \centering
    \includegraphics[width=0.90\linewidth]{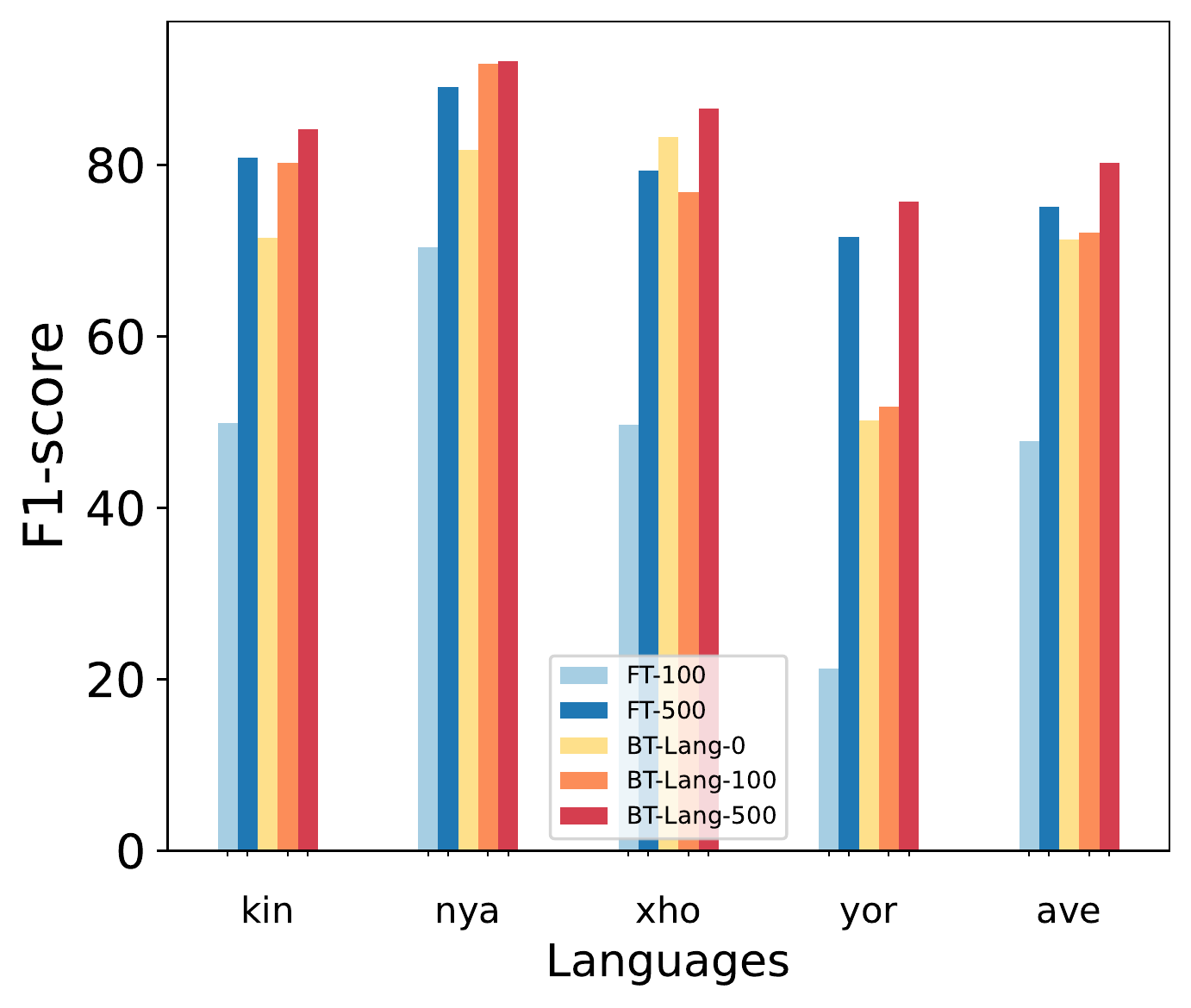}
    \vspace{-3mm}
    \caption{\textbf{Sample Efficiency Results} for 100 and 500 samples in the target language, model fine-tuned from a PLM (e.g. FT-100 -- trained on 100 samples from the target language) or fine-tuned from the best transfer language NER model (e.g BT-Lang-0 -- trained on 0 samples from the target language or zero-shot)}
    \label{fig:transfer_image_sample_eff}
    \vspace{-3mm}
    
\end{figure}

\subsection{Sample Efficiency Results}

\autoref{fig:transfer_image_sample_eff} shows the performance when the model is trained on a few target language samples compared to when the best transfer language is used prior to fine-tuning on the same number of target language samples. We show the results for four languages (which reflect common patterns across all languages) and an average (\texttt{ave}) over the 20 languages. As seen in the figure, models achieve less than $50$ F1 when we train on 100 sentences and over $75$ F1 when training on 500 sentences. In practice, annotating 100 sentences takes about 30 minutes while annotating 500 sentences takes around 2 hours and 30 minutes; therefore, slightly more annotation effort can yield a substantial quality improvement. We also find that using the best transfer language in zero-shot settings gives a performance very close to annotating 500 samples in most cases, showing the importance of transfer language selection. By additionally fine-tuning the model on 100 or 500 target language samples, we can further improve the NER performance. \autoref{sec:sample_efficiency_resu} provides the sample efficiency results for individual languages.

\section{Conclusion}

In this paper, we present the creation of MasakhaNER 2.0, the largest NER dataset for 20 diverse African languages and provide strong baseline results on the corpus by fine-tuning multilingual PLMs on in-language NER and multilingual datasets.
Additionally, we analyze cross-lingual transfer in an Africa-centric setting, showing the importance of choosing the best transfer language in both zero-shot and few-shot scenarios.
Using English as the default transfer language can have detrimental effects, and choosing a more appropriate language substantially improves fine-tuned NER models.
By analyzing data-dependent, geographical, and typological features for transfer in NER, we conclude that geographical distance and entity overlap contribute most effectively to transfer performance.

\section*{Acknowledgements}
This work was carried out with support from Lacuna Fund, an initiative co-founded by The Rockefeller Foundation, Google.org, and Canada’s International Development Research Centre. David Adelani acknowledges the EU-funded Horizon 2020 projects: COMPRISE under grant agreement No. 3081705 and ROXANNE under grant number 833635. Vukosi Marivate acknowledges funding from ABSA covering the ABSA Data Science chair as well as funding from the Google Research Scholar program. We thank Mengzhou Xia, Antonis Anastasopoulos, and Fahim Faisal for their help with the LangRank code and data geography analysis. We thank Haneul Yoo for her comment on the initial Korean NER result. We thank Heng Ji and Ying Lin for
providing the ELISA NER tool used for annotation. We thank Daniel D'souza for helping to set-up ELISA NER annotation tool. We thank Kelechi Ogueji for providing monolingual corpus for isiZulu. We thank Google for providing GCP credits to run some of the experiments. We thank the ML Group, Luleå University of Technology, for the compute resources for running some of the experiments. Finally, we thank the Masakhane leadership, 
Melissa Omino, Davor Orlič and Knowledge4All for their administrative support throughout the project. 

\section*{Limitations}

\paragraph{Some Language families not covered}
While we try to cover 20 topologically diverse languages and language families, a few locations in Africa and smaller language family groups were not covered. For example, languages from the Khoisan and Austronesian (like Malagasy) family were not covered. Also, languages spoken in the central Africa like South Sudan, Chad, and DRC were not covered. 

\paragraph{News Domain Data}
As the data we annotated belonged to the news domain, models trained from this data may not generalize well to other domains.
In particular, the models may not perform well on more casual text that may use different vocabulary, discuss different entities, and contain more orthographic variation. This limitation also applies for the English NER Corpus. 

\paragraph{Generalizability of Transfer Learning Findings}
As we only experimented with one task (NER), our findings regarding effective approaches to transfer learning for African languages and PLMs may not generalize to other tasks (e.g. machine translation, part of speech tagging); other features of language similarity may be more important for other tasks.

\paragraph{Explaining Transfer Learning Findings}
We found that the \texttt{LangRank} model could not predict the top transfer languages with 100\% accuracy.
This suggests that there are other, unknown factors that could affect transfer performance, which we did not explore.
For example, there is still work to be done to understand the sociolinguistic connections and language contact conditions that may correlate with effective transfer.




\section*{Ethics Statement}

Our research process has been deeply rooted in the principles of participatory AI research~\cite{nekoto_etal_2020_participatory}, where the populations most affected by the research---the native speakers of the languages in this case---are involved throughout the project as stakeholders.

We believe our work will be of benefit to the speakers of the included languages by enabling better language technology for their languages.
By keeping them engaged throughout the process and as collaborators in this work, we have been able to become aware of any potential harms.
As the data we use for annotation is news data that was already publicly available, the release of our annotation is unlikely to cause unintended harm.

However, there are always potential unintended consequences when creating NER data and models.
The data selection, annotation, adjudication, and model training process can all introduce biases that may have negative effects.
Specifically, within each language, the models trained may perform better when processing names that commonly appear in newswire, and worse when processing names belonging to entities less well-represented in the news domain, propagating biases to downstream tasks.



\bibliography{anthology,custom}

\begin{thebibliography}{80}
\expandafter\ifx\csname natexlab\endcsname\relax\def\natexlab#1{#1}\fi

\bibitem[{Abdulmumin et~al.(2022)Abdulmumin, Dash, Dawud, Parida, Muhammad,
  Ahmad, Panda, Bojar, Galadanci, and Bello}]{abdulmumin-etal-2022-hausa}
Idris Abdulmumin, Satya~Ranjan Dash, Musa~Abdullahi Dawud, Shantipriya Parida,
  Shamsuddeen Muhammad, Ibrahim~Sa{'}id Ahmad, Subhadarshi Panda, Ond{\v{r}}ej
  Bojar, Bashir~Shehu Galadanci, and Bello~Shehu Bello. 2022.
\newblock \href {https://aclanthology.org/2022.lrec-1.694} {{H}ausa visual
  genome: A dataset for multi-modal {E}nglish to {H}ausa machine translation}.
\newblock In \emph{Proceedings of the Thirteenth Language Resources and
  Evaluation Conference}, pages 6471--6479, Marseille, France. European
  Language Resources Association.

\bibitem[{Adelani et~al.(2021{\natexlab{a}})Adelani, Ruiter, Alabi, Adebonojo,
  Ayeni, Adeyemi, Awokoya, and Espa{\~n}a-Bonet}]{adelani-etal-2021-effect}
David Adelani, Dana Ruiter, Jesujoba Alabi, Damilola Adebonojo, Adesina Ayeni,
  Mofe Adeyemi, Ayodele~Esther Awokoya, and Cristina Espa{\~n}a-Bonet.
  2021{\natexlab{a}}.
\newblock \href {https://aclanthology.org/2021.mtsummit-research.6} {The effect
  of domain and diacritics in {Y}oruba{--}{E}nglish neural machine
  translation}.
\newblock In \emph{Proceedings of Machine Translation Summit XVIII: Research
  Track}, pages 61--75, Virtual. Association for Machine Translation in the
  Americas.

\bibitem[{Adelani et~al.(2021{\natexlab{b}})Adelani, Abbott, Neubig, D{'}souza,
  Kreutzer, Lignos, Palen-Michel, Buzaaba, Rijhwani, Ruder, Mayhew, Azime,
  Muhammad, Emezue, Nakatumba-Nabende, Ogayo, Anuoluwapo, Gitau, Mbaye, Alabi,
  Yimam, Gwadabe, Ezeani, Niyongabo, Mukiibi, Otiende, Orife, David, Ngom,
  Adewumi, Rayson, Adeyemi, Muriuki, Anebi, Chukwuneke, Odu, Wairagala,
  Oyerinde, Siro, Bateesa, Oloyede, Wambui, Akinode, Nabagereka, Katusiime,
  Awokoya, MBOUP, Gebreyohannes, Tilaye, Nwaike, Wolde, Faye, Sibanda, Ahia,
  Dossou, Ogueji, DIOP, Diallo, Akinfaderin, Marengereke, and
  Osei}]{adelani-etal-2021-masakhaner}
David~Ifeoluwa Adelani, Jade Abbott, Graham Neubig, Daniel D{'}souza, Julia
  Kreutzer, Constantine Lignos, Chester Palen-Michel, Happy Buzaaba, Shruti
  Rijhwani, Sebastian Ruder, Stephen Mayhew, Israel~Abebe Azime, Shamsuddeen~H.
  Muhammad, Chris~Chinenye Emezue, Joyce Nakatumba-Nabende, Perez Ogayo, Aremu
  Anuoluwapo, Catherine Gitau, Derguene Mbaye, Jesujoba Alabi, Seid~Muhie
  Yimam, Tajuddeen~Rabiu Gwadabe, Ignatius Ezeani, Rubungo~Andre Niyongabo,
  Jonathan Mukiibi, Verrah Otiende, Iroro Orife, Davis David, Samba Ngom, Tosin
  Adewumi, Paul Rayson, Mofetoluwa Adeyemi, Gerald Muriuki, Emmanuel Anebi,
  Chiamaka Chukwuneke, Nkiruka Odu, Eric~Peter Wairagala, Samuel Oyerinde,
  Clemencia Siro, Tobius~Saul Bateesa, Temilola Oloyede, Yvonne Wambui, Victor
  Akinode, Deborah Nabagereka, Maurice Katusiime, Ayodele Awokoya, Mouhamadane
  MBOUP, Dibora Gebreyohannes, Henok Tilaye, Kelechi Nwaike, Degaga Wolde,
  Abdoulaye Faye, Blessing Sibanda, Orevaoghene Ahia, Bonaventure F.~P. Dossou,
  Kelechi Ogueji, Thierno~Ibrahima DIOP, Abdoulaye Diallo, Adewale Akinfaderin,
  Tendai Marengereke, and Salomey Osei. 2021{\natexlab{b}}.
\newblock \href {https://doi.org/10.1162/tacl_a_00416} {{M}asakha{NER}: Named
  entity recognition for {A}frican languages}.
\newblock \emph{Transactions of the Association for Computational Linguistics},
  9:1116--1131.

\bibitem[{Adelani et~al.(2022)Adelani, Alabi, Fan, Kreutzer, Shen, Reid,
  Ruiter, Klakow, Nabende, Chang, Gwadabe, Sackey, Dossou, Emezue, Leong,
  Beukman, Muhammad, Jarso, Yousuf, Rubungo, HACHEME, Wairagala, Nasir,
  Ajibade, Ajayi, Gitau, Abbott, Ahmed, Ochieng, Aremu, Ogayo, Mukiibi, Kabore,
  KALIPE, Mbaye, Tapo, Koagne, Munkoh-Buabeng, Wagner, Abdulmumin, Awokoya,
  Buzaaba, Sibanda, Bukula, and Manthalu}]{adelani_mafand}
David~Ifeoluwa Adelani, Jesujoba~Oluwadara Alabi, Angela Fan, Julia Kreutzer,
  Xiaoyu Shen, Machel Reid, Dana Ruiter, Dietrich Klakow, Peter Nabende, Ernie
  Chang, Tajuddeen Gwadabe, Freshia Sackey, Bonaventure F.~P. Dossou,
  Chris~Chinenye Emezue, Colin Leong, Michael Beukman, Shamsuddeen~Hassan
  Muhammad, Guyo~Dub Jarso, Oreen Yousuf, Andre~Niyongabo Rubungo, Gilles
  HACHEME, Eric~Peter Wairagala, Muhammad~Umair Nasir, Benjamin~Ayoade Ajibade,
  Oluwaseyi~Ajayi Ajayi, Yvonne~Wambui Gitau, Jade Abbott, Mohamed Ahmed,
  Millicent Ochieng, Anuoluwapo Aremu, Perez Ogayo, Jonathan Mukiibi,
  Fatoumata~Ouoba Kabore, Godson~Koffi KALIPE, Derguene Mbaye,
  Allahsera~Auguste Tapo, Victoire~Memdjokam Koagne, Edwin Munkoh-Buabeng,
  Valencia Wagner, Idris Abdulmumin, Ayodele Awokoya, Happy Buzaaba, Blessing
  Sibanda, Andiswa Bukula, and Sam Manthalu. 2022.
\newblock \href {https://openreview.net/forum?id=EtZ9h4Lqs5-} {A few thousand
  translations go a long way! leveraging pre-trained models for african news
  translation}.
\newblock In \emph{NAACL-HLT}.

\bibitem[{Ahuja et~al.(2022)Ahuja, Kumar, Dandapat, and
  Choudhury}]{ahuja-etal-2022-multi}
Kabir Ahuja, Shanu Kumar, Sandipan Dandapat, and Monojit Choudhury. 2022.
\newblock \href {https://aclanthology.org/2022.acl-long.374} {Multi task
  learning for zero shot performance prediction of multilingual models}.
\newblock In \emph{Proceedings of the 60th Annual Meeting of the Association
  for Computational Linguistics (Volume 1: Long Papers)}, pages 5454--5467,
  Dublin, Ireland. Association for Computational Linguistics.

\bibitem[{Alabi et~al.(2020)Alabi, Amponsah-Kaakyire, Adelani, and
  Espa{\~n}a-Bonet}]{alabi-etal-2020-massive}
Jesujoba Alabi, Kwabena Amponsah-Kaakyire, David Adelani, and Cristina
  Espa{\~n}a-Bonet. 2020.
\newblock \href {https://aclanthology.org/2020.lrec-1.335} {Massive vs. curated
  embeddings for low-resourced languages: the case of {Y}or{\`u}b{\'a} and
  {T}wi}.
\newblock In \emph{Proceedings of the 12th Language Resources and Evaluation
  Conference}, pages 2754--2762, Marseille, France. European Language Resources
  Association.

\bibitem[{Alabi et~al.(2022)Alabi, Adelani, Mosbach, and
  Klakow}]{alabi-etal-2022-adapting}
Jesujoba~O. Alabi, David~Ifeoluwa Adelani, Marius Mosbach, and Dietrich Klakow.
  2022.
\newblock \href {https://aclanthology.org/2022.coling-1.382} {Adapting
  pre-trained language models to {A}frican languages via multilingual adaptive
  fine-tuning}.
\newblock In \emph{Proceedings of the 29th International Conference on
  Computational Linguistics}, pages 4336--4349, Gyeongju, Republic of Korea.
  International Committee on Computational Linguistics.

\bibitem[{Babou and Loporcaro(2016)}]{BabouLoporcaro+2016+1+57}
Cheikh~Anta Babou and Michele Loporcaro. 2016.
\newblock \href {https://doi.org/doi:10.1515/jall-2016-0001} {Noun classes and
  grammatical gender in wolof}.
\newblock \emph{Journal of African Languages and Linguistics}, 37(1):1--57.

\bibitem[{Benajiba et~al.(2007)Benajiba, Rosso, and
  Bened{\'i}Ruiz}]{benajiba_arabic_ner}
Yassine Benajiba, Paolo Rosso, and Jos{\'e}~Miguel Bened{\'i}Ruiz. 2007.
\newblock Anersys: An arabic named entity recognition system based on maximum
  entropy.
\newblock In \emph{Computational Linguistics and Intelligent Text Processing},
  pages 143--153, Berlin, Heidelberg. Springer Berlin Heidelberg.

\bibitem[{Beukman(2022)}]{beukman2021analysing}
Michael Beukman. 2022.
\newblock \href {https://openreview.net/forum?id=HKWMFqfN8b5} {Analysing the
  effects of transfer learning on low-resourced named entity recognition
  performance}.
\newblock In \emph{3rd Workshop on African Natural Language Processing}.

\bibitem[{Bodomo and Marfo(2002)}]{Bodomo2002TheMO}
Adams Bodomo and Charles Marfo. 2002.
\newblock The morphophonology of noun classes in dagaare and akan.

\bibitem[{Chung et~al.(2021)Chung, Fevry, Tsai, Johnson, and
  Ruder}]{chung2021rethinking}
Hyung~Won Chung, Thibault Fevry, Henry Tsai, Melvin Johnson, and Sebastian
  Ruder. 2021.
\newblock \href {https://openreview.net/forum?id=xpFFI_NtgpW} {Rethinking
  embedding coupling in pre-trained language models}.
\newblock In \emph{International Conference on Learning Representations}.

\bibitem[{Clark et~al.(2020)Clark, Luong, Le, and Manning}]{clark2020electra}
Kevin Clark, Minh-Thang Luong, Quoc~V. Le, and Christopher~D. Manning. 2020.
\newblock \href {https://openreview.net/pdf?id=r1xMH1BtvB} {{ELECTRA}:
  Pre-training text encoders as discriminators rather than generators}.
\newblock In \emph{ICLR}.

\bibitem[{Conneau et~al.(2020)Conneau, Khandelwal, Goyal, Chaudhary, Wenzek,
  Guzm{\'a}n, Grave, Ott, Zettlemoyer, and
  Stoyanov}]{conneau-etal-2020-unsupervised}
Alexis Conneau, Kartikay Khandelwal, Naman Goyal, Vishrav Chaudhary, Guillaume
  Wenzek, Francisco Guzm{\'a}n, Edouard Grave, Myle Ott, Luke Zettlemoyer, and
  Veselin Stoyanov. 2020.
\newblock \href {https://doi.org/10.18653/v1/2020.acl-main.747} {Unsupervised
  cross-lingual representation learning at scale}.
\newblock In \emph{Proceedings of the 58th Annual Meeting of the Association
  for Computational Linguistics}, pages 8440--8451, Online. Association for
  Computational Linguistics.

\bibitem[{De~Cao et~al.(2022)De~Cao, Wu, Popat, Artetxe, Goyal, Plekhanov,
  Zettlemoyer, Cancedda, Riedel, and Petroni}]{de-cao-etal-2022-multilingual}
Nicola De~Cao, Ledell Wu, Kashyap Popat, Mikel Artetxe, Naman Goyal, Mikhail
  Plekhanov, Luke Zettlemoyer, Nicola Cancedda, Sebastian Riedel, and Fabio
  Petroni. 2022.
\newblock \href {https://doi.org/10.1162/tacl_a_00460} {Multilingual
  autoregressive entity linking}.
\newblock \emph{Transactions of the Association for Computational Linguistics},
  10:274--290.

\bibitem[{de~Vries et~al.(2022)de~Vries, Wieling, and
  Nissim}]{de-vries-etal-2022-make}
Wietse de~Vries, Martijn Wieling, and Malvina Nissim. 2022.
\newblock \href {https://aclanthology.org/2022.acl-long.529} {Make the best of
  cross-lingual transfer: Evidence from {POS} tagging with over 100 languages}.
\newblock In \emph{Proceedings of the 60th Annual Meeting of the Association
  for Computational Linguistics (Volume 1: Long Papers)}, pages 7676--7685,
  Dublin, Ireland. Association for Computational Linguistics.

\bibitem[{De~Waal et~al.(2006)De~Waal, Louis, and Venter}]{De_Waal2006-dh}
A~J De~Waal, A~L Louis, and J~P Venter. 2006.
\newblock Named entity recognition in a south african context.

\bibitem[{Devlin et~al.(2019{\natexlab{a}})Devlin, Chang, Lee, and
  Toutanova}]{devlin-etal-2019-bert}
Jacob Devlin, Ming-Wei Chang, Kenton Lee, and Kristina Toutanova.
  2019{\natexlab{a}}.
\newblock \href {https://doi.org/10.18653/v1/N19-1423} {{BERT}: Pre-training of
  deep bidirectional transformers for language understanding}.
\newblock In \emph{Proceedings of the 2019 Conference of the North {A}merican
  Chapter of the Association for Computational Linguistics: Human Language
  Technologies, Volume 1 (Long and Short Papers)}, pages 4171--4186,
  Minneapolis, Minnesota. Association for Computational Linguistics.

\bibitem[{Devlin et~al.(2019{\natexlab{b}})Devlin, Chang, Lee, and
  Toutanova}]{bert}
Jacob Devlin, Ming-Wei Chang, Kenton Lee, and Kristina Toutanova.
  2019{\natexlab{b}}.
\newblock \href {https://doi.org/10.18653/v1/N19-1423} {{BERT}: Pre-training of
  deep bidirectional transformers for language understanding}.
\newblock In \emph{Proceedings of the 2019 Conference of the North {A}merican
  Chapter of the Association for Computational Linguistics: Human Language
  Technologies, Volume 1 (Long and Short Papers)}, Minneapolis, Minnesota.
  Association for Computational Linguistics.

\bibitem[{Dryer and Haspelmath(2013)}]{wals}
Matthew~S. Dryer and Martin Haspelmath, editors. 2013.
\newblock \href {https://wals.info/} {\emph{WALS Online}}.
\newblock Max Planck Institute for Evolutionary Anthropology, Leipzig.

\bibitem[{Dumitrescu and Avram(2020)}]{dumitrescu-avram-2020-introducing}
Stefan~Daniel Dumitrescu and Andrei-Marius Avram. 2020.
\newblock \href {https://aclanthology.org/2020.lrec-1.546} {Introducing {RONEC}
  - the {R}omanian named entity corpus}.
\newblock In \emph{Proceedings of the 12th Language Resources and Evaluation
  Conference}, pages 4436--4443, Marseille, France. European Language Resources
  Association.

\bibitem[{Ebrahimi et~al.(2022)Ebrahimi, Mager, Oncevay, Chaudhary, Chiruzzo,
  Fan, Ortega, Ramos, Rios, Meza~Ruiz, Gim{\'e}nez-Lugo, Mager, Neubig, Palmer,
  Coto-Solano, Vu, and Kann}]{ebrahimi-etal-2022-americasnli}
Abteen Ebrahimi, Manuel Mager, Arturo Oncevay, Vishrav Chaudhary, Luis
  Chiruzzo, Angela Fan, John Ortega, Ricardo Ramos, Annette Rios, Ivan~Vladimir
  Meza~Ruiz, Gustavo Gim{\'e}nez-Lugo, Elisabeth Mager, Graham Neubig, Alexis
  Palmer, Rolando Coto-Solano, Thang Vu, and Katharina Kann. 2022.
\newblock \href {https://aclanthology.org/2022.acl-long.435} {{A}mericas{NLI}:
  Evaluating zero-shot natural language understanding of pretrained
  multilingual models in truly low-resource languages}.
\newblock In \emph{Proceedings of the 60th Annual Meeting of the Association
  for Computational Linguistics (Volume 1: Long Papers)}, pages 6279--6299,
  Dublin, Ireland. Association for Computational Linguistics.

\bibitem[{Eiselen(2016)}]{eiselen-2016-government}
Roald Eiselen. 2016.
\newblock \href {https://aclanthology.org/L16-1533} {Government domain named
  entity recognition for {S}outh {A}frican languages}.
\newblock In \emph{Proceedings of the Tenth International Conference on
  Language Resources and Evaluation ({LREC}'16)}, pages 3344--3348,
  Portoro{\v{z}}, Slovenia. European Language Resources Association (ELRA).

\bibitem[{Faisal et~al.(2022)Faisal, Wang, and
  Anastasopoulos}]{faisal-etal-2022-dataset}
Fahim Faisal, Yinkai Wang, and Antonios Anastasopoulos. 2022.
\newblock \href {https://aclanthology.org/2022.acl-long.239} {Dataset
  geography: Mapping language data to language users}.
\newblock In \emph{Proceedings of the 60th Annual Meeting of the Association
  for Computational Linguistics (Volume 1: Long Papers)}, pages 3381--3411,
  Dublin, Ireland. Association for Computational Linguistics.

\bibitem[{$\forall$ et~al.(2020)$\forall$, Nekoto, Marivate, Matsila, Fasubaa,
  Fagbohungbe, Akinola, Muhammad, Kabongo~Kabenamualu, Osei, Sackey, Niyongabo,
  Macharm, Ogayo, Ahia, Berhe, Adeyemi, Mokgesi-Selinga, Okegbemi, Martinus,
  Tajudeen, Degila, Ogueji, Siminyu, Kreutzer, Webster, Ali, Abbott, Orife,
  Ezeani, Dangana, Kamper, Elsahar, Duru, Kioko, Espoir, van Biljon, Whitenack,
  Onyefuluchi, Emezue, Dossou, Sibanda, Bassey, Olabiyi, Ramkilowan, {\"O}ktem,
  Akinfaderin, and Bashir}]{nekoto_etal_2020_participatory}
{}~$\forall$, Wilhelmina Nekoto, Vukosi Marivate, Tshinondiwa Matsila, Timi
  Fasubaa, Taiwo Fagbohungbe, Solomon~Oluwole Akinola, Shamsuddeen Muhammad,
  Salomon Kabongo~Kabenamualu, Salomey Osei, Freshia Sackey, Rubungo~Andre
  Niyongabo, Ricky Macharm, Perez Ogayo, Orevaoghene Ahia, Musie~Meressa Berhe,
  Mofetoluwa Adeyemi, Masabata Mokgesi-Selinga, Lawrence Okegbemi, Laura
  Martinus, Kolawole Tajudeen, Kevin Degila, Kelechi Ogueji, Kathleen Siminyu,
  Julia Kreutzer, Jason Webster, Jamiil~Toure Ali, Jade Abbott, Iroro Orife,
  Ignatius Ezeani, Idris~Abdulkadir Dangana, Herman Kamper, Hady Elsahar,
  Goodness Duru, Ghollah Kioko, Murhabazi Espoir, Elan van Biljon, Daniel
  Whitenack, Christopher Onyefuluchi, Chris~Chinenye Emezue, Bonaventure F.~P.
  Dossou, Blessing Sibanda, Blessing Bassey, Ayodele Olabiyi, Arshath
  Ramkilowan, Alp {\"O}ktem, Adewale Akinfaderin, and Abdallah Bashir. 2020.
\newblock \href {https://www.aclweb.org/anthology/2020.findings-emnlp.195}
  {Participatory research for low-resourced machine translation: A case study
  in {A}frican languages}.
\newblock In \emph{Findings of the Association for Computational Linguistics:
  EMNLP 2020}, Online.

\bibitem[{Freitas et~al.(2010)Freitas, Mota, Santos, Oliveira, and
  Carvalho}]{freitas-etal-2010-second}
Cl{\'a}udia Freitas, Cristina Mota, Diana Santos, Hugo~Gon{\c{c}}alo Oliveira,
  and Paula Carvalho. 2010.
\newblock \href
  {http://www.lrec-conf.org/proceedings/lrec2010/pdf/412_Paper.pdf} {Second
  {HAREM}: Advancing the state of the art of named entity recognition in
  {P}ortuguese}.
\newblock In \emph{Proceedings of the Seventh International Conference on
  Language Resources and Evaluation ({LREC}'10)}, Valletta, Malta. European
  Language Resources Association (ELRA).

\bibitem[{Gruzitis et~al.(2018)Gruzitis, Pretkalnina, Saulite, Rituma,
  Nespore-Berzkalne, Znotins, and Paikens}]{gruzitis-etal-2018-creation}
Normunds Gruzitis, Lauma Pretkalnina, Baiba Saulite, Laura Rituma, Gunta
  Nespore-Berzkalne, Arturs Znotins, and Peteris Paikens. 2018.
\newblock \href {https://aclanthology.org/L18-1714} {Creation of a balanced
  state-of-the-art multilayer corpus for {NLU}}.
\newblock In \emph{Proceedings of the Eleventh International Conference on
  Language Resources and Evaluation ({LREC} 2018)}, Miyazaki, Japan. European
  Language Resources Association (ELRA).

\bibitem[{Hammarstr{\"o}m et~al.(2018)Hammarstr{\"o}m, Forkel, and
  Haspelmath}]{hammarstrom2018glottolog}
Harald Hammarstr{\"o}m, Robert Forkel, and Martin Haspelmath. 2018.
\newblock Glottolog 3.0.
\newblock \emph{Max Planck Institute for the Science of Human History}.

\bibitem[{He et~al.(2021)He, Gao, and Chen}]{He2021DeBERTaV3ID}
Pengcheng He, Jianfeng Gao, and Weizhu Chen. 2021.
\newblock Debertav3: Improving deberta using electra-style pre-training with
  gradient-disentangled embedding sharing.
\newblock \emph{ArXiv}, abs/2111.09543.

\bibitem[{Hedderich et~al.(2020)Hedderich, Adelani, Zhu, Alabi, Markus, and
  Klakow}]{hedderich-etal-2020-transfer}
Michael~A. Hedderich, David Adelani, Dawei Zhu, Jesujoba Alabi, Udia Markus,
  and Dietrich Klakow. 2020.
\newblock \href {https://doi.org/10.18653/v1/2020.emnlp-main.204} {Transfer
  learning and distant supervision for multilingual transformer models: A study
  on {A}frican languages}.
\newblock In \emph{Proceedings of the 2020 Conference on Empirical Methods in
  Natural Language Processing (EMNLP)}, pages 2580--2591, Online. Association
  for Computational Linguistics.

\bibitem[{Hu et~al.(2020)Hu, Ruder, Siddhant, Neubig, Firat, and
  Johnson}]{pmlr-v119-hu20b}
Junjie Hu, Sebastian Ruder, Aditya Siddhant, Graham Neubig, Orhan Firat, and
  Melvin Johnson. 2020.
\newblock \href {https://proceedings.mlr.press/v119/hu20b.html} {{XTREME}: A
  massively multilingual multi-task benchmark for evaluating cross-lingual
  generalisation}.
\newblock In \emph{Proceedings of the 37th International Conference on Machine
  Learning}, volume 119 of \emph{Proceedings of Machine Learning Research},
  pages 4411--4421. PMLR.

\bibitem[{Hvingelby et~al.(2020)Hvingelby, Pauli, Barrett, Rosted, Lidegaard,
  and S{\o}gaard}]{hvingelby-etal-2020-dane}
Rasmus Hvingelby, Amalie~Brogaard Pauli, Maria Barrett, Christina Rosted,
  Lasse~Malm Lidegaard, and Anders S{\o}gaard. 2020.
\newblock \href {https://aclanthology.org/2020.lrec-1.565} {{D}a{NE}: A named
  entity resource for {D}anish}.
\newblock In \emph{Proceedings of the 12th Language Resources and Evaluation
  Conference}, pages 4597--4604, Marseille, France. European Language Resources
  Association.

\bibitem[{Jibril and Tantug(2022)}]{Jibril2022ANECAA}
Ebrahim~Chekol Jibril and A.~C{\"u}neyd Tantug. 2022.
\newblock Anec: An amharic named entity corpus and transformer based
  recognizer.
\newblock \emph{ArXiv}, abs/2207.00785.

\bibitem[{Johansen(2019)}]{johansen2019ner}
Bjarte Johansen. 2019.
\newblock Named-entity recognition for norwegian.
\newblock In \emph{Proceedings of the 22nd Nordic Conference on Computational
  Linguistics, NoDaLiDa}.

\bibitem[{Junior et~al.(2015)Junior, Macedo, Bispo, Oliveira, Silva, and
  Barbosa}]{c2015paramopama}
C.~Junior, H.~Macedo, T.~Bispo, F.~Oliveira, N.~Silva, and L.~Barbosa. 2015.
\newblock Paramopama: a brazilian-portuguese corpus for named entity
  recognition.
\newblock In \emph{12th National Meeting on Artificial and Computational
  Intelligence (ENIAC)}.

\bibitem[{K et~al.(2020)K, Wang, Mayhew, and Roth}]{k2020Crosslingual}
Karthikeyan K, Zihan Wang, Stephen Mayhew, and Dan Roth. 2020.
\newblock \href {https://openreview.net/forum?id=HJeT3yrtDr} {Cross-lingual
  ability of multilingual bert: An empirical study}.
\newblock In \emph{International Conference on Learning Representations}.

\bibitem[{Karamolegkou and
  Stymne(2021)}]{karamolegkou-stymne-2021-investigation}
Antonia Karamolegkou and Sara Stymne. 2021.
\newblock \href {https://aclanthology.org/2021.nodalida-main.32} {Investigation
  of transfer languages for parsing {L}atin: Italic branch vs. {H}ellenic
  branch}.
\newblock In \emph{Proceedings of the 23rd Nordic Conference on Computational
  Linguistics (NoDaLiDa)}, pages 315--320, Reykjavik, Iceland (Online).
  Link{\"o}ping University Electronic Press, Sweden.

\bibitem[{Khairunnisa et~al.(2020)Khairunnisa, Imankulova, and
  Komachi}]{khairunnisa-etal-2020-towards}
Siti~Oryza Khairunnisa, Aizhan Imankulova, and Mamoru Komachi. 2020.
\newblock \href {https://aclanthology.org/2020.aacl-srw.10} {Towards a
  standardized dataset on {I}ndonesian named entity recognition}.
\newblock In \emph{Proceedings of the 1st Conference of the Asia-Pacific
  Chapter of the Association for Computational Linguistics and the 10th
  International Joint Conference on Natural Language Processing: Student
  Research Workshop}, pages 64--71, Suzhou, China. Association for
  Computational Linguistics.

\bibitem[{Konoshenko and Shavarina(2019)}]{Konoshenko2019AMS}
Maria~Yu Konoshenko and Dasha Shavarina. 2019.
\newblock A microtypological survey of noun classes in kwa.
\newblock \emph{Journal of African Languages and Linguistics}, 40:114 -- 75.

\bibitem[{Lauscher et~al.(2020)Lauscher, Ravishankar, Vuli{\'c}, and
  Glava{\v{s}}}]{lauscher-etal-2020-zero}
Anne Lauscher, Vinit Ravishankar, Ivan Vuli{\'c}, and Goran Glava{\v{s}}. 2020.
\newblock \href {https://doi.org/10.18653/v1/2020.emnlp-main.363} {From zero to
  hero: {O}n the limitations of zero-shot language transfer with multilingual
  {T}ransformers}.
\newblock In \emph{Proceedings of the 2020 Conference on Empirical Methods in
  Natural Language Processing (EMNLP)}, pages 4483--4499, Online. Association
  for Computational Linguistics.

\bibitem[{Lewis(2009)}]{lewis2009ethnologue}
M~Paul Lewis. 2009.
\newblock \href {http://www.ethnologue.com/16/} {\emph{Ethnologue: Languages of
  the world Sixteenth Edition}}.
\newblock SIL international.

\bibitem[{Lhoest et~al.(2021)Lhoest, Villanova~del Moral, Jernite, Thakur, von
  Platen, Patil, Chaumond, Drame, Plu, Tunstall, Davison, {\v{S}}a{\v{s}}ko,
  Chhablani, Malik, Brandeis, Le~Scao, Sanh, Xu, Patry, McMillan-Major, Schmid,
  Gugger, Delangue, Matussi{\`e}re, Debut, Bekman, Cistac, Goehringer, Mustar,
  Lagunas, Rush, and Wolf}]{lhoest-etal-2021-datasets}
Quentin Lhoest, Albert Villanova~del Moral, Yacine Jernite, Abhishek Thakur,
  Patrick von Platen, Suraj Patil, Julien Chaumond, Mariama Drame, Julien Plu,
  Lewis Tunstall, Joe Davison, Mario {\v{S}}a{\v{s}}ko, Gunjan Chhablani,
  Bhavitvya Malik, Simon Brandeis, Teven Le~Scao, Victor Sanh, Canwen Xu,
  Nicolas Patry, Angelina McMillan-Major, Philipp Schmid, Sylvain Gugger,
  Cl{\'e}ment Delangue, Th{\'e}o Matussi{\`e}re, Lysandre Debut, Stas Bekman,
  Pierric Cistac, Thibault Goehringer, Victor Mustar, Fran{\c{c}}ois Lagunas,
  Alexander Rush, and Thomas Wolf. 2021.
\newblock \href {https://doi.org/10.18653/v1/2021.emnlp-demo.21} {Datasets: A
  community library for natural language processing}.
\newblock In \emph{Proceedings of the 2021 Conference on Empirical Methods in
  Natural Language Processing: System Demonstrations}, pages 175--184, Online
  and Punta Cana, Dominican Republic. Association for Computational
  Linguistics.

\bibitem[{Lin et~al.(2018)Lin, Costello, Zhang, Lu, Ji, Mayfield, and
  McNamee}]{lin-etal-2018-platforms}
Ying Lin, Cash Costello, Boliang Zhang, Di~Lu, Heng Ji, James Mayfield, and
  Paul McNamee. 2018.
\newblock \href {https://doi.org/10.18653/v1/P18-4001} {Platforms for
  non-speakers annotating names in any language}.
\newblock In \emph{Proceedings of {ACL} 2018, System Demonstrations}, pages
  1--6, Melbourne, Australia. Association for Computational Linguistics.

\bibitem[{Lin et~al.(2019)Lin, Chen, Lee, Li, Zhang, Xia, Rijhwani, He, Zhang,
  Ma, Anastasopoulos, Littell, and Neubig}]{lin-etal-2019-choosing}
Yu-Hsiang Lin, Chian-Yu Chen, Jean Lee, Zirui Li, Yuyan Zhang, Mengzhou Xia,
  Shruti Rijhwani, Junxian He, Zhisong Zhang, Xuezhe Ma, Antonios
  Anastasopoulos, Patrick Littell, and Graham Neubig. 2019.
\newblock \href {https://doi.org/10.18653/v1/P19-1301} {Choosing transfer
  languages for cross-lingual learning}.
\newblock In \emph{Proceedings of the 57th Annual Meeting of the Association
  for Computational Linguistics}, pages 3125--3135, Florence, Italy.
  Association for Computational Linguistics.

\bibitem[{Littell et~al.(2017)Littell, Mortensen, Lin, Kairis, Turner, and
  Levin}]{littell-etal-2017-uriel}
Patrick Littell, David~R. Mortensen, Ke~Lin, Katherine Kairis, Carlisle Turner,
  and Lori Levin. 2017.
\newblock \href {https://aclanthology.org/E17-2002} {{URIEL} and lang2vec:
  Representing languages as typological, geographical, and phylogenetic
  vectors}.
\newblock In \emph{Proceedings of the 15th Conference of the {E}uropean Chapter
  of the Association for Computational Linguistics: Volume 2, Short Papers},
  pages 8--14, Valencia, Spain. Association for Computational Linguistics.

\bibitem[{Liu et~al.(2021)Liu, Fu, Xiao, Yuan, Chang, Dai, Liu, Ye, and
  Neubig}]{liu-etal-2021-explainaboard}
Pengfei Liu, Jinlan Fu, Yang Xiao, Weizhe Yuan, Shuaichen Chang, Junqi Dai,
  Yixin Liu, Zihuiwen Ye, and Graham Neubig. 2021.
\newblock \href {https://doi.org/10.18653/v1/2021.acl-demo.34}
  {{E}xplaina{B}oard: An explainable leaderboard for {NLP}}.
\newblock In \emph{Proceedings of the 59th Annual Meeting of the Association
  for Computational Linguistics and the 11th International Joint Conference on
  Natural Language Processing: System Demonstrations}, pages 280--289, Online.
  Association for Computational Linguistics.

\bibitem[{Magnini et~al.(2008)Magnini, Cappelli, Tamburini, Bosco, Mazzei,
  Lombardo, Bertagna, Calzolari, Toral, Bartalesi~Lenzi, Sprugnoli, and
  Speranza}]{magnini-etal-2008-evaluation}
Bernardo Magnini, Amedeo Cappelli, Fabio Tamburini, Cristina Bosco, Alessandro
  Mazzei, Vincenzo Lombardo, Francesca Bertagna, Nicoletta Calzolari, Antonio
  Toral, Valentina Bartalesi~Lenzi, Rachele Sprugnoli, and Manuela Speranza.
  2008.
\newblock \href
  {http://www.lrec-conf.org/proceedings/lrec2008/pdf/630_paper.pdf} {Evaluation
  of natural language tools for {I}talian: {EVALITA} 2007}.
\newblock In \emph{Proceedings of the Sixth International Conference on
  Language Resources and Evaluation ({LREC}'08)}, Marrakech, Morocco. European
  Language Resources Association (ELRA).

\bibitem[{Mayhew et~al.(2019)Mayhew, Tsygankova, and
  Roth}]{mayhew-etal-2019-ner}
Stephen Mayhew, Tatiana Tsygankova, and Dan Roth. 2019.
\newblock \href {https://doi.org/10.18653/v1/D19-1650} {{ner and pos when
  nothing is capitalized}}.
\newblock In \emph{Proceedings of the 2019 Conference on Empirical Methods in
  Natural Language Processing and the 9th International Joint Conference on
  Natural Language Processing (EMNLP-IJCNLP)}, pages 6256--6261, Hong Kong,
  China. Association for Computational Linguistics.

\bibitem[{Melzian(1933)}]{melzian_1933}
Hans~J. Melzian. 1933.
\newblock \href {https://doi.org/10.1017/S0041977X00105828} {Introduction to
  the phonology of the bantu languages}.
\newblock \emph{Bulletin of the School of Oriental and African Studies},
  7(1):246–247.

\bibitem[{Moran et~al.(2014)Moran, McCloy, and Wright}]{moran2014phoible}
Steven Moran, D~McCloy, and R~Wright. 2014.
\newblock Phoible online. max planck institute for evolutionary anthropology,
  leipzig.

\bibitem[{Muhammad et~al.(2022)Muhammad, Adelani, Ruder, Ahmad, Abdulmumin,
  Bello, Choudhury, Emezue, Abdullahi, Aremu, Jorge, and
  Brazdil}]{Muhammad2022NaijaSentiAN}
Shamsuddeen~Hassan Muhammad, David~Ifeoluwa Adelani, Sebastian Ruder,
  Ibrahim~Sa{'}id Ahmad, Idris Abdulmumin, Bello~Shehu Bello, Monojit
  Choudhury, Chris~Chinenye Emezue, Saheed~Salahudeen Abdullahi, Anuoluwapo
  Aremu, Al{\'\i}pio Jorge, and Pavel Brazdil. 2022.
\newblock \href {https://aclanthology.org/2022.lrec-1.63} {{N}aija{S}enti: A
  nigerian {T}witter sentiment corpus for multilingual sentiment analysis}.
\newblock In \emph{Proceedings of the Thirteenth Language Resources and
  Evaluation Conference}, pages 590--602, Marseille, France. European Language
  Resources Association.

\bibitem[{Neudecker(2016)}]{neudecker-2016-open}
Clemens Neudecker. 2016.
\newblock \href {https://aclanthology.org/L16-1689} {An open corpus for named
  entity recognition in historic newspapers}.
\newblock In \emph{Proceedings of the Tenth International Conference on
  Language Resources and Evaluation ({LREC}'16)}, pages 4348--4352,
  Portoro{\v{z}}, Slovenia. European Language Resources Association (ELRA).

\bibitem[{Nichols and Bickel(2013)}]{wals-59}
Johanna Nichols and Balthasar Bickel. 2013.
\newblock \href {https://wals.info/chapter/59} {Possessive classification}.
\newblock In Matthew~S. Dryer and Martin Haspelmath, editors, \emph{The World
  Atlas of Language Structures Online}. Max Planck Institute for Evolutionary
  Anthropology, Leipzig.

\bibitem[{Nurse and Philippson(2006)}]{Van_de_Velde2006-fz}
Derek Nurse and Gerard Philippson, editors. 2006.
\newblock \emph{The Bantu Languages}.
\newblock Routledge Language Family Series. Routledge, London, England.

\bibitem[{Obeid et~al.(2020)Obeid, Zalmout, Khalifa, Taji, Oudah, Alhafni,
  Inoue, Eryani, Erdmann, and Habash}]{obeid-etal-2020-camel}
Ossama Obeid, Nasser Zalmout, Salam Khalifa, Dima Taji, Mai Oudah, Bashar
  Alhafni, Go~Inoue, Fadhl Eryani, Alexander Erdmann, and Nizar Habash. 2020.
\newblock \href {https://aclanthology.org/2020.lrec-1.868} {{CAM}e{L} tools: An
  open source python toolkit for {A}rabic natural language processing}.
\newblock In \emph{Proceedings of the 12th Language Resources and Evaluation
  Conference}, pages 7022--7032, Marseille, France. European Language Resources
  Association.

\bibitem[{Ogueji et~al.(2021)Ogueji, Zhu, and Lin}]{ogueji-etal-2021-small}
Kelechi Ogueji, Yuxin Zhu, and Jimmy Lin. 2021.
\newblock \href {https://doi.org/10.18653/v1/2021.mrl-1.11} {Small data? no
  problem! exploring the viability of pretrained multilingual language models
  for low-resourced languages}.
\newblock In \emph{Proceedings of the 1st Workshop on Multilingual
  Representation Learning}, pages 116--126, Punta Cana, Dominican Republic.
  Association for Computational Linguistics.

\bibitem[{Oladipo et~al.(2022)Oladipo, Ogundepo, Ogueji, and
  Lin}]{oladipo2022an}
Akintunde Oladipo, Odunayo Ogundepo, Kelechi Ogueji, and Jimmy Lin. 2022.
\newblock \href {https://openreview.net/forum?id=HOZmF9MV8Wc} {An exploration
  of vocabulary size and transfer effects in multilingual language models for
  african languages}.
\newblock In \emph{3rd Workshop on African Natural Language Processing}.

\bibitem[{Oosthuysen(2016)}]{oosthuysen_xhosa}
JC~Oosthuysen. 2016.
\newblock \href {http://www.jstor.org/stable/j.ctv1nzg1tj} {\emph{The Grammar
  of isiXhosa}}, 1 edition.
\newblock African Sun Media.

\bibitem[{Park et~al.(2021)Park, Moon, Kim, Cho, Han, Park, Song, Kim, Song,
  Oh, Lee, Oh, Lyu, Jeong, Lee, Seo, Lee, Kim, Lee, Jang, Do, Kim, Lim, Lee,
  Park, Shin, Kim, Park, Oh, Ha, and Cho}]{park2021klue}
Sungjoon Park, Jihyung Moon, Sungdong Kim, Won~Ik Cho, Jiyoon Han, Jangwon
  Park, Chisung Song, Junseong Kim, Yongsook Song, Taehwan Oh, Joohong Lee,
  Juhyun Oh, Sungwon Lyu, Younghoon Jeong, Inkwon Lee, Sangwoo Seo, Dongjun
  Lee, Hyunwoo Kim, Myeonghwa Lee, Seongbo Jang, Seungwon Do, Sunkyoung Kim,
  Kyungtae Lim, Jongwon Lee, Kyumin Park, Jamin Shin, Seonghyun Kim, Lucy Park,
  Alice Oh, Jungwoo Ha, and Kyunghyun Cho. 2021.
\newblock \href {http://arxiv.org/abs/2105.09680} {Klue: Korean language
  understanding evaluation}.

\bibitem[{Payne et~al.(2017)Payne, Pacchiarotti, and Bosire}]{Payne2017}
Doris~L. Payne, Sara Pacchiarotti, and Mokaya Bosire, editors. 2017.
\newblock \emph{Diversity in {African} languages}.
\newblock Number~1 in Contemporary African Linguistics. Language Science Press,
  Berlin.

\bibitem[{Pfeiffer et~al.(2020)Pfeiffer, Vuli{\'c}, Gurevych, and
  Ruder}]{pfeiffer-etal-2020-mad}
Jonas Pfeiffer, Ivan Vuli{\'c}, Iryna Gurevych, and Sebastian Ruder. 2020.
\newblock \href {https://doi.org/10.18653/v1/2020.emnlp-main.617} {{MAD-X}:
  {A}n {A}dapter-{B}ased {F}ramework for {M}ulti-{T}ask {C}ross-{L}ingual
  {T}ransfer}.
\newblock In \emph{Proceedings of the 2020 Conference on Empirical Methods in
  Natural Language Processing (EMNLP)}, pages 7654--7673, Online. Association
  for Computational Linguistics.

\bibitem[{Pires et~al.(2019)Pires, Schlinger, and
  Garrette}]{pires-etal-2019-multilingual}
Telmo Pires, Eva Schlinger, and Dan Garrette. 2019.
\newblock \href {https://doi.org/10.18653/v1/P19-1493} {How multilingual is
  multilingual {BERT}?}
\newblock In \emph{Proceedings of the 57th Annual Meeting of the Association
  for Computational Linguistics}, pages 4996--5001, Florence, Italy.
  Association for Computational Linguistics.

\bibitem[{Ponti et~al.(2020)Ponti, Glava{\v{s}}, Majewska, Liu, Vuli{\'c}, and
  Korhonen}]{ponti-etal-2020-xcopa}
Edoardo~Maria Ponti, Goran Glava{\v{s}}, Olga Majewska, Qianchu Liu, Ivan
  Vuli{\'c}, and Anna Korhonen. 2020.
\newblock \href {https://doi.org/10.18653/v1/2020.emnlp-main.185} {{XCOPA}: A
  multilingual dataset for causal commonsense reasoning}.
\newblock In \emph{Proceedings of the 2020 Conference on Empirical Methods in
  Natural Language Processing (EMNLP)}, pages 2362--2376, Online. Association
  for Computational Linguistics.

\bibitem[{Poostchi et~al.(2016)Poostchi, Zare~Borzeshi, Abdous, and
  Piccardi}]{poostchi-etal-2016-personer}
Hanieh Poostchi, Ehsan Zare~Borzeshi, Mohammad Abdous, and Massimo Piccardi.
  2016.
\newblock \href {https://aclanthology.org/C16-1319} {{P}erso{NER}: {P}ersian
  named-entity recognition}.
\newblock In \emph{Proceedings of {COLING} 2016, the 26th International
  Conference on Computational Linguistics: Technical Papers}, pages 3381--3389,
  Osaka, Japan. The COLING 2016 Organizing Committee.

\bibitem[{Priatama et~al.(2022)Priatama, , and Setiawan}]{Priatama2022-tv}
A~R Priatama, , and Y~Setiawan. 2022.
\newblock Regression models for estimating aboveground biomass and stand volume
  using landsat-based indices in post-mining area.
\newblock \emph{J. Manaj. Hutan Trop. (J. Trop. For. Manag.)}, 28(1):1--14.

\bibitem[{Pruksachatkun et~al.(2020)Pruksachatkun, Phang, Liu, Htut, Zhang,
  Pang, Vania, Kann, and Bowman}]{pruksachatkun-etal-2020-intermediate}
Yada Pruksachatkun, Jason Phang, Haokun Liu, Phu~Mon Htut, Xiaoyi Zhang,
  Richard~Yuanzhe Pang, Clara Vania, Katharina Kann, and Samuel~R. Bowman.
  2020.
\newblock \href {https://doi.org/10.18653/v1/2020.acl-main.467}
  {Intermediate-task transfer learning with pretrained language models: When
  and why does it work?}
\newblock In \emph{Proceedings of the 58th Annual Meeting of the Association
  for Computational Linguistics}, pages 5231--5247, Online. Association for
  Computational Linguistics.

\bibitem[{Reid et~al.(2021)Reid, Hu, Neubig, and
  Matsuo}]{reid-etal-2021-afromt}
Machel Reid, Junjie Hu, Graham Neubig, and Yutaka Matsuo. 2021.
\newblock \href {https://doi.org/10.18653/v1/2021.emnlp-main.99} {{A}fro{MT}:
  Pretraining strategies and reproducible benchmarks for translation of 8
  {A}frican languages}.
\newblock In \emph{Proceedings of the 2021 Conference on Empirical Methods in
  Natural Language Processing}, pages 1306--1320, Online and Punta Cana,
  Dominican Republic. Association for Computational Linguistics.

\bibitem[{Ruder et~al.(2021)Ruder, Constant, Botha, Siddhant, Firat, Fu, Liu,
  Hu, Garrette, Neubig, and Johnson}]{ruder-etal-2021-xtreme}
Sebastian Ruder, Noah Constant, Jan Botha, Aditya Siddhant, Orhan Firat, Jinlan
  Fu, Pengfei Liu, Junjie Hu, Dan Garrette, Graham Neubig, and Melvin Johnson.
  2021.
\newblock \href {https://doi.org/10.18653/v1/2021.emnlp-main.802}
  {{XTREME}-{R}: Towards more challenging and nuanced multilingual evaluation}.
\newblock In \emph{Proceedings of the 2021 Conference on Empirical Methods in
  Natural Language Processing}, pages 10215--10245, Online and Punta Cana,
  Dominican Republic. Association for Computational Linguistics.

\bibitem[{Ruokolainen et~al.(2019)Ruokolainen, Kauppinen, Silfverberg, and
  Lind{\'e}n}]{ruokolainen2019finnish}
Teemu Ruokolainen, Pekka Kauppinen, Miikka Silfverberg, and Krister Lind{\'e}n.
  2019.
\newblock A finnish news corpus for named entity recognition.
\newblock \emph{Language Resources and Evaluation}, pages 1--26.

\bibitem[{{Singh} et~al.(2019){Singh}, {Padia}, and {Joshi}}]{singh_nepali_ner}
O.~M. {Singh}, A.~{Padia}, and A.~{Joshi}. 2019.
\newblock \href {https://doi.org/10.1109/CIC48465.2019.00031} {Named entity
  recognition for nepali language}.
\newblock In \emph{2019 IEEE 5th International Conference on Collaboration and
  Internet Computing (CIC)}, pages 184--190.

\bibitem[{Strassel and Tracey(2016)}]{strassel-tracey-2016-lorelei}
Stephanie Strassel and Jennifer Tracey. 2016.
\newblock \href {https://aclanthology.org/L16-1521} {{LORELEI} language packs:
  Data, tools, and resources for technology development in low resource
  languages}.
\newblock In \emph{Proceedings of the Tenth International Conference on
  Language Resources and Evaluation ({LREC}'16)}, pages 3273--3280,
  Portoro{\v{z}}, Slovenia. European Language Resources Association (ELRA).

\bibitem[{Szarvas et~al.(2006)Szarvas, Farkas, Felf{\"o}ldi, Kocsor, and
  Csirik}]{szarvas-etal-2006-highly}
Gy{\"o}rgy Szarvas, Rich{\'a}rd Farkas, L{\'a}szl{\'o} Felf{\"o}ldi, Andr{\'a}s
  Kocsor, and J{\'a}nos Csirik. 2006.
\newblock \href {http://www.lrec-conf.org/proceedings/lrec2006/pdf/365_pdf.pdf}
  {A highly accurate named entity corpus for {H}ungarian}.
\newblock In \emph{Proceedings of the Fifth International Conference on
  Language Resources and Evaluation ({LREC}{'}06)}, Genoa, Italy. European
  Language Resources Association (ELRA).

\bibitem[{Tjong Kim~Sang(2002)}]{tjong-kim-sang-2002-introduction}
Erik~F. Tjong Kim~Sang. 2002.
\newblock \href {https://aclanthology.org/W02-2024} {Introduction to the
  {C}o{NLL}-2002 shared task: Language-independent named entity recognition}.
\newblock In \emph{{COLING}-02: The 6th Conference on Natural Language Learning
  2002 ({C}o{NLL}-2002)}.

\bibitem[{Tjong Kim~Sang and
  De~Meulder(2003)}]{tjong-kim-sang-de-meulder-2003-introduction}
Erik~F. Tjong Kim~Sang and Fien De~Meulder. 2003.
\newblock \href {https://aclanthology.org/W03-0419} {Introduction to the
  {C}o{NLL}-2003 shared task: Language-independent named entity recognition}.
\newblock In \emph{Proceedings of the Seventh Conference on Natural Language
  Learning at {HLT}-{NAACL} 2003}, pages 142--147.

\bibitem[{Versteegh(2001)}]{arabic_swahili_influence}
Kees Versteegh. 2001.
\newblock \href {https://doi.org/10.1163/157005801323163825} {Linguistic
  contacts between arabic and other languages}.
\newblock \emph{Arabica}, 48:470--508.

\bibitem[{Wolf et~al.(2020)Wolf, Debut, Sanh, Chaumond, Delangue, Moi, Cistac,
  Rault, Louf, Funtowicz, Davison, Shleifer, von Platen, Ma, Jernite, Plu, Xu,
  Le~Scao, Gugger, Drame, Lhoest, and Rush}]{wolf-etal-2020-transformers}
Thomas Wolf, Lysandre Debut, Victor Sanh, Julien Chaumond, Clement Delangue,
  Anthony Moi, Pierric Cistac, Tim Rault, Remi Louf, Morgan Funtowicz, Joe
  Davison, Sam Shleifer, Patrick von Platen, Clara Ma, Yacine Jernite, Julien
  Plu, Canwen Xu, Teven Le~Scao, Sylvain Gugger, Mariama Drame, Quentin Lhoest,
  and Alexander Rush. 2020.
\newblock \href {https://doi.org/10.18653/v1/2020.emnlp-demos.6} {Transformers:
  State-of-the-art natural language processing}.
\newblock In \emph{Proceedings of the 2020 Conference on Empirical Methods in
  Natural Language Processing: System Demonstrations}, pages 38--45, Online.
  Association for Computational Linguistics.

\bibitem[{Wu and Dredze(2019)}]{wu-dredze-2019-beto}
Shijie Wu and Mark Dredze. 2019.
\newblock \href {https://doi.org/10.18653/v1/D19-1077} {Beto, bentz, becas: The
  surprising cross-lingual effectiveness of {BERT}}.
\newblock In \emph{Proceedings of the 2019 Conference on Empirical Methods in
  Natural Language Processing and the 9th International Joint Conference on
  Natural Language Processing (EMNLP-IJCNLP)}, pages 833--844, Hong Kong,
  China. Association for Computational Linguistics.

\bibitem[{Xia et~al.(2020)Xia, Anastasopoulos, Xu, Yang, and
  Neubig}]{xia-etal-2020-predicting}
Mengzhou Xia, Antonios Anastasopoulos, Ruochen Xu, Yiming Yang, and Graham
  Neubig. 2020.
\newblock \href {https://doi.org/10.18653/v1/2020.acl-main.764} {Predicting
  performance for natural language processing tasks}.
\newblock In \emph{Proceedings of the 58th Annual Meeting of the Association
  for Computational Linguistics}, pages 8625--8646, Online. Association for
  Computational Linguistics.

\bibitem[{Yimam et~al.(2020)Yimam, Alemayehu, Ayele, and
  Biemann}]{yimam-etal-2020-exploring}
Seid~Muhie Yimam, Hizkiel~Mitiku Alemayehu, Abinew Ayele, and Chris Biemann.
  2020.
\newblock \href {https://doi.org/10.18653/v1/2020.coling-main.91} {Exploring
  {A}mharic sentiment analysis from social media texts: Building annotation
  tools and classification models}.
\newblock In \emph{Proceedings of the 28th International Conference on
  Computational Linguistics}, pages 1048--1060, Barcelona, Spain (Online).
  International Committee on Computational Linguistics.

\bibitem[{Yohannes and Amagasa(2022)}]{tigrinya_ner}
Hailemariam~Mehari Yohannes and Toshiyuki Amagasa. 2022.
\newblock \href {https://doi.org/10.1145/3477314.3507066} {Named-entity
  recognition for a low-resource language using pre-trained language model}.
\newblock In \emph{Proceedings of the 37th ACM/SIGAPP Symposium on Applied
  Computing}, SAC '22, page 837–844, New York, NY, USA. Association for
  Computing Machinery.

\end{thebibliography}
\bibliographystyle{acl_natbib}

\appendix

\section{Data Source and Splits}
\label{sec:appendix_data_source}
\autoref{tab:data_stat_entities_split} shows the MasakhaNER 2.0 language, data source, train/dev/test split, and the number of tokens per entity type.

\begin{table*}[th!]
 \footnotesize
 \begin{center}
 \resizebox{\textwidth}{!}{%
  \begin{tabular}{lllrrrrrr}
    \toprule
     &  \multicolumn{2}{c}{\textbf{}} & \multicolumn{4}{c}{\textbf{\# Tokens}} & \textbf{\% Entities}   & \\ \cmidrule(lr){4-7}
    \textbf{Language} & \textbf{Data Source} & \textbf{Train / dev / test} & \textbf{PER} & \textbf{LOC} & \textbf{ORG} & \textbf{DATE} & \textbf{in Tokens}   & \textbf{\#Tokens} \\
    \midrule
    Bambara (\texttt{bam}) & \mafand~\cite{adelani_mafand} & 4462/ 638/ 1274 & 4281 & 2557 & 429 & 2898 & 6.5 & 155,552\\
    \ghomala (\texttt{bbj}) & \mafand~\cite{adelani_mafand} & 3384/ 483/ 966 & 2464 & 1371 & 1586 & 2457 & 11.3 & 69,474 \\
    \ewe (\texttt{ewe})  & \mafand~\cite{adelani_mafand} & 3505/ 501/ 1001 & 3931 & 5168 & 2064 & 2665 & 15.3 & 90420  \\
    Fon (\texttt{fon}) & \mafand~\cite{adelani_mafand}  & 4343/ 621/ 1240  & 3572 & 2595 & 3082 & 5120 &  8.3 & 173,099\\
    Hausa (\texttt{hau}) & Kano Focus and Freedom Radio & 5716/ 816/ 1633  & 9853 & 6759 & 7089 & 7251 & 14.0 & 221,086\\
    Igbo (\texttt{ibo})  & IgboRadio and Ka \d{O}d\d{I} Taa  & 7634/ 1090/ 2181  & 8532 & 7077 & 5418 & 4727 &  7.5 & 344,095\\
    Kinyarwanda (\texttt{kin}) & IGIHE, Rwanda & 7825/ 1118/ 2235  & 6889 & 8960 & 7012 & 8187 & 12.6 & 245,933 \\
    Luganda (\texttt{lug}) & \mafand~\cite{adelani_mafand} & 4942/ 706/ 1412  & 6058 & 3706 & 5441 & 3484 & 15.6 & 120,119\\
    Luo (\texttt{luo}) & \mafand~\cite{adelani_mafand} & 5161/ 737/ 1474  & 6806 & 5605 & 7099 & 7359 & 11.7 & 229,927\\
    Mossi (\texttt{mos}) & \mafand~\cite{adelani_mafand} & 4532/ 648/ 1294  & 2804 & 3044 & 3209 & 6334 & 9.2 & 168,141\\
    Naija (\texttt{pcm}) & \mafand~\cite{adelani_mafand} & 5646/ 806/ 1613  & 4711 & 5077 & 5940 & 3654 & 9.4 & 206,404\\
    Chichewa (\texttt{nya}) & Nation Online Malawi & 6250/ 893/ 1785  & 9657 & 4600 & 5924 & 4308 & 9.3 & 263,622\\
    Shona (\texttt{sna}) & VOA Shona & 6207/ 887/ 1773  & 10667 & 5289 & 12418 & 3423 &  16.2 & 195,834\\
    Swahili (\texttt{swa}) & VOA Swahili & 6593/ 942/ 1883 & 9510 & 10515 & 6515 & 5331 &  12.7 & 251,678\\
    Setswana (\texttt{tsn}) & \mafand~\cite{adelani_mafand} & 3489/ 499/ 996 & 3991 & 2285 & 2905 & 3190 & 8.8 & 141,069\\
    Akan/Twi (\texttt{twi}) & \mafand~\cite{adelani_mafand} & 4240/ 605/ 1211  & 3588 & 2474 & 2375 & 1433 & 6.3 & 155,985\\
    Wolof (\texttt{wol}) & \mafand~\cite{adelani_mafand} & 4593/ 656/ 1312  & 3588 & 2474 & 2375 & 1433 & 7.4 & 181,048\\
    \xhosa (\texttt{xho}) & Isolezwe Newspaper & 5718/ 817/ 1633  & 8098 & 3087 & 5633 & 2433 & 15.1 & 127,222\\
    \yoruba (\texttt{yor}) & Voice of Nigeria and Asejere & 6877/ 983/ 1964  & 8537 & 5819 & 6998 & 6372 & 11.4 & 244,144\\
    \zulu (\texttt{zul}) & Isolezwe Newspaper & 5848/ 836/ 1670 & 5050 & 1900 & 5229 & 2012 & 11.0 & 128,658\\
    \bottomrule
  \end{tabular}
  }
  \vspace{-3mm}
  \caption{\textbf{Languages and Data Splits for MasakhaNER 2.0 Corpus}. Distribution of the number of entities}
  \vspace{-4mm}
  \label{tab:data_stat_entities_split}
  \end{center}
\end{table*}

\label{sec:appendix_lang_char}
 \begin{table*}[t]
 \footnotesize
 \begin{center}
 \resizebox{\textwidth}{!}{%
  \begin{tabular}{lrlp{50mm}llllll}
    \toprule
     & \textbf{No. of} &\textbf{Latin Letters} &  &  &  &  & \textbf{Morphological} & \textbf{Inflectional} & \textbf{Noun}\\
    \textbf{Language} & \textbf{Letters} & \textbf{Omitted} & \textbf{Letters added}  & \textbf{Tonality} & \textbf{diacritics} & \textbf{Word Order} & \textbf{typology} & \textbf{Morphology (WALS)} & \textbf{Classes} \\
    \midrule
    Bambara (bam) & 27 & q,v,x & \textepsilon, \textopeno, \textltailn, \textipa{\ng} &  yes, 2 tones & yes & SVO \& SOV & isolating & strong suffixing & absent \\
    \ghomala (bbj) & 40 & q, w, x, y & bv, dz, \textschwa, a\textschwa, \textepsilon, gh, ny, nt, \textipa{\ng}, \textipa{\ng}k, \textopeno, pf, mpf, sh, ts, \textbaru, zh, '  &  yes, 5 tones & yes & SVO & agglutinative & strong prefixing & active, 6 \\
    \ewe (ewe) & 35  & c, j, q & \textrtaild, dz, \textepsilon, \textflorin, gb, \textgamma, kp, ny, \textipa{\ng}, \textopeno, ts, \textscriptv   & yes, 3 tones  & yes & SVO & isolating & equal prefixing and suffixing & vestigial \\
    Fon (fon) & 33 & q & \textrtaild, \textepsilon,gb, hw, kp, ny, \textopeno, xw & yes, 3 tones & yes  & SVO & isolating & little affixation & vestigial  \\
    Hausa (hau) & 44 & p,q,v,x & \texthtb, \texthtd, \texthtk, \begin{tfour}\m{y}\end{tfour}, kw, {\texthtk}w, gw, ky, {\texthtk}y, gy, sh, ts  & yes, 2 tones  & no & SVO & agglutinative & little affixation & absent \\
    Igbo (ibo) & 34 & c, q, x & ch, gb, gh, gw, kp, kw, nw, ny, {\d o}, \.{o}, sh, {\d u}& yes, 2 tones  & yes & SVO & agglutinative & little affixation & vestigial  \\
    Kinyarwanda (kin) & 30 & q, x & cy, jy, nk, nt, ny, sh & yes, 2 tones & no & SVO & agglutinative & strong prefixing & active, 16 \\
    Luganda (lug) & 25 & h, q, x & \textipa{\ng}, ny & yes, 3 tones & no & SVO & agglutinative & strong prefixing & active, 20 \\
    Luo (luo) & 31 & c, q, x, v, z &   ch, dh, mb, nd, ng’, ng, ny, nj, th, sh & yes, 4 tones & no & SVO  & agglutinative & equal prefixing and suffixing & absent \\
    Mossi (mos) & 26 & c, j, q, x  & ', \textepsilon, \textiota, \textscriptv & yes, 2 tones  & yes  & SVO  & isolating & strongly suffixing & active, 11 \\
    Chichewa (nya) & 31 & q, x, y & ch, kh, ng, \textipa{\ng}, ph, tch, th, \^{w} & yes, 2 tones & no  & SVO  & agglutinative & strong prefixing & active, 17\\
    Naija (pcm) & 26 & -- & -- & no  & no & SVO & mostly analytic & strongly suffixing & absent\\
    Shona (sna) & 29 & c, l, q, x & bh, ch, dh, nh, sh, vh, zh  &  yes, 2 tones & no & SVO & agglutinative & strong prefixing & active, 20 \\
    Swahili (swa) & 33 & x, q & ch, dh, gh, kh, ng', ny, sh, th, ts & no & yes   & SVO  & agglutinative & strong suffixing & active, 18 \\
    Setswana (tsn) & 36 & c, q, v, x, z & \^{e}, kg, kh, ng, ny, \^o, ph, \v{s}, th, tl, tlh, ts, tsh, t\v{s}, t\v{s}h & yes, 2 tones & no & SVO & agglutinative & strong prefixing & active, 18\\
    Akan/Twi (twi) & 22 & c,j,q,v,x,z & \textepsilon, \textopeno & yes, 5 tones  & no & SVO & isolating & strong prefixing &  active, 6\\
    Wolof (wol) &29& h,v,z & \textipa{\ng}, \`a, \'e, \"{e}, \'o, \~{n}  & no & yes  & SVO & agglutinative & strong suffixing & active, 10 \\
    \xhosa (xho) & 68 & -- & bh, ch, dl, dy, dz, gc, gq, gr, gx, hh, hl, kh, kr, lh, mh, ng, ngc, ngh, ngq, ngx, nkq, nkx, nh, nkc, nx, ny, nyh, ph, qh, rh, sh, th, ths, thsh, ts, tsh, ty, tyh, wh, xh, yh, zh &  yes, 2 tones & no & SVO & agglutinative & strong prefixing & active, 17\\
    \yoruba (yor) & 25 & c, q, v, x, z & {\d e}, gb, {\d s} , {\d o} & yes, 3 tones & yes  & SVO & isolating & little affixation & vestigial, 2 \\
    \zulu (zul) & 55 & -- & nx, ts, nq, ph, hh, ny, gq, hl, bh, nj, ch, ngc, ngq, th, ngx, kl, ntsh, sh, kh, tsh, ng, nk, gx, xh, gc, mb, dl, nc, qh &  yes, 3 tones & no & SVO  & agglutinative & strong prefixing & active, 17 \\
    \bottomrule
  \end{tabular}
  }
  \vspace{-3mm}
  \caption{Linguistic Characteristics of the Languages}
  \label{tab:lang_char}
  \end{center}
\end{table*}

\section{Language Characteristics}
\label{sec:lang_charateristics}
\autoref{tab:lang_char} provides the details about the language characteristics. 

\subsection{Morphology and Noun classes} 
Many African languages are morphologically rich. According to the World Atlas of Language Structures~\cite[WALS;][]{wals-59}, 16 of our languages employ strong prefixing or suffixing inflections. Niger-Congo languages are known for their system of noun classification. 12 of the languages \textit{actively} make use of between 6--20 noun classes, including all Bantu languages and \ghomala, Mossi, Akan and Wolof~\cite{Van_de_Velde2006-fz,Payne2017,Bodomo2002TheMO,BabouLoporcaro+2016+1+57}. While noun classes are often marked using affixes on the head word in Bantu languages, some non-Bantu languages, e.g., Wolof make use of a dependent such as a determiner that is not attached to the head word. 
For the other Niger-Congo languages such as Fon, Ewe, Igbo and \yoruba, the use of noun classes is merely \textit{vestigial}~\cite{Konoshenko2019AMS}. For example, \yoruba only distinguishes between human and non-human nouns. Bambara is the only Niger-Congo language without noun classes, and some have argued that the Mande family should be regarded as an independent language family.  
Three of our languages from the Southern Bantu family (\shona, \xhosa and \zulu) capitalize proper names after the noun class prefix as in the language names themselves. This characteristic limits the transfer learning of NER from languages without this feature, since NER models overfit on capitalization~\cite{mayhew-etal-2019-ner}. 

\begin{table*}[th!]
 \footnotesize
 \begin{center}
  \begin{tabular}{llrrr}
    \toprule
    \textbf{Language} & \textbf{Data Source} & \textbf{\# Train} & \textbf{\# dev} & \textbf{\# test} \\
    \midrule
    Amharic (\texttt{amh}) & MasakhaNER 1.0~\cite{adelani-etal-2021-masakhaner} & 1,750 & 250 & 500 \\
    Arabic (\texttt{ara}) & ANERcorp~\cite{benajiba_arabic_ner,obeid-etal-2020-camel} & 3,472 & 500 & 924 \\
    Danish (\texttt{dan})  & DANE~\cite{hvingelby-etal-2020-dane} & 4,383 & 564 & 565 \\
    German (\texttt{deu}) & CoNLL03~\cite{tjong-kim-sang-de-meulder-2003-introduction}  & 12,152 & 2,867 & 3,005  \\
    English (\texttt{eng}) & CoNLL03~\cite{tjong-kim-sang-de-meulder-2003-introduction} & 14,041 & 3,250 & 3,453  \\
    Spanish (\texttt{spa})  & CoNLL02~\cite{tjong-kim-sang-2002-introduction}  & 8,322 & 1,914 & 1,516  \\
    Farsi (\texttt{fas}) & PersoNER~\cite{poostchi-etal-2016-personer} & 4,121 & 1,000 & 2,560  \\
    Finnish (\texttt{fin}) & FINER~\cite{ruokolainen2019finnish} & 13,497 & 986 & 3,512  \\
    French (\texttt{fra}) & Europeana~\cite{neudecker-2016-open} & 9,546 & 2,045 & 2,047  \\
    Hungarian (\texttt{hun}) & Hungarian MTI~\cite{szarvas-etal-2006-highly} & 4,532 & 648 & 1,294  \\
    Indonesia (\texttt{ind}) & \cite{khairunnisa-etal-2020-towards} & 6,707 & 1,437 & 1,438  \\
    Italian (\texttt{ita}) & I-CAB EVALITA 2007 \& 2009~\cite{magnini-etal-2008-evaluation} & 11,227 & 4,136 & 2,068  \\
    Korean (\texttt{kor}) & KLUE~\cite{park2021klue} & 20,008 & 1,000 & 5,000 \\
    Latvian (\texttt{lav}) & \cite{gruzitis-etal-2018-creation} & 7,997 & 1,713 & 1,715 \\
    Nepali (\texttt{nep}) & \cite{singh_nepali_ner} & 2,301 & 328 & 659 \\
    Dutch (\texttt{nld}) & CoNLL02~\cite{tjong-kim-sang-2002-introduction} & 15,806 & 2,895 & 5,195  \\
    Norwegian (\texttt{nor}) & \cite{johansen2019ner} & 15,696 & 2,410 & 1,939 \\
    Portuguese (\texttt{por}) & Second HAREM~\cite{freitas-etal-2010-second} \& Paramopama~\cite{c2015paramopama}  & 11,258 & 2,412 & 2,414  \\
    Romanian (\texttt{ron}) & RONEC~\cite{dumitrescu-avram-2020-introducing} &  5,886 & 1,000 & 2,453 \\
    Swedish (\texttt{swe}) & ``swedish\textunderscore ner\textunderscore corpus'' on HuggingFace Datasets~\cite{lhoest-etal-2021-datasets} & 9,000 & 1,330 & 2,000 \\
    Ukrainian (\texttt{ukr}) & ``benjamin/ner-uk'' on HuggingFace Datasets~\cite{lhoest-etal-2021-datasets} & 10,833 & 1,307 & 668 \\
    Chinese (\texttt{zho}) & ``msra\textunderscore ner'' on HuggingFace Datasets~\cite{lhoest-etal-2021-datasets} & 45,057 & 3,442 & 1,721 \\
    \bottomrule
  \end{tabular}
  \vspace{-3mm}
  \caption{\textbf{Languages and Data Splits for Other NER Datasets}. }
  \vspace{-4mm}
  \label{tab:non_african_data_split}
  \end{center}
\end{table*}

\begin{table*}[ht]
 \begin{center}
 \resizebox{\textwidth}{!}{
   \begin{tabular}{lrrrrrrrrrrrrrrr}
    \toprule
\multirow{2}{*}{\textbf{Language}} & \multicolumn{5}{c}{\textbf{XLM-R-large}} & \multicolumn{5}{c}{\textbf{mDeBERTaV3-base}} & \multicolumn{5}{c}{\textbf{AfroXLMR-large}} \\
    \cmidrule(lr){2-6} \cmidrule(lr){7-11} \cmidrule(lr){12-16}
& all & 0-freq &  $\Delta$ 0-freq & long & $\Delta$ long  & all & 0-freq &  $\Delta$ 0-freq & long & $\Delta$ long & all & 0-freq &  $\Delta$ 0-freq & long & $\Delta$ long \\
    \midrule
\texttt{bam} & 79.4 & 62.3 & -17.1 & 74.7 & -4.7 & 81.3 & 66.3 & -15.0 & 78.6 & -2.7 & 82.1 & 67.2 & -14.9 & 81.1 & -1.0 \\
\texttt{bbj} & 74.8 & 66.1 & -8.7 & 87.4 & 12.6 & 75.0 & 65.8 & -9.2 & 63.9 & -11.1 & 76.5 & 65.8 & -10.7 & 80.0 & 3.5 \\
\texttt{ewe} & 89.5 & 75.6 & -13.9 & 70.6 & -18.9 & 90.0 & 76.9 & -13.1 & 70 & -20.0 & 91.0 & 79.7 & -11.3 & 74.2 & -16.8 \\
\texttt{fon} & 81.5 & 71.2 & -10.3 & 69.6 & -11.9 & 83.3 & 74.5 & -8.8 & 68.1 & -15.2 & 82.8 & 73.6 & -9.2 & 68.7 & -14.1 \\
\texttt{hau} & 87.4 & 83.8 & -3.6 & 77.6 & -9.8 & 84.8 & 80.0 & -4.8 & 72.2 & -12.6 & 87.8 & 84.6 & -3.2 & 78.1 & -9.7 \\
\texttt{ibo} & 87.0 & 77.4 & -9.6 & 75.6 & -11.4 & 89.7 & 82.6 & -7.1 & 71.8 & -17.9 & 89.1 & 80.9 & -8.2 & 64.0 & -25.1 \\
\texttt{kin} & 84.1 & 74.9 & -9.2 & 75.3 & -8.8 & 86.2 & 79.0 & -7.2 & 75.3 & -10.9 & 87.8 & 81.7 & -6.1 & 77.1 & -10.7 \\
\texttt{lug} & 87.3 & 75.3 & -12.0 & 74.1 & -13.2 & 88.7 & 77.4 & -11.3 & 78.6 & -10.1 & 89.4 & 79.7 & -9.7 & 74.7 & -14.7 \\
\texttt{mos} & 77.1 & 69.5 & -7.6 & 55.8 & -21.3 & 78.0 & 71.2 & -6.8 & 58.9 & -19.1 & 77.5 & 70.2 & -7.3 & 60.1 & -17.4 \\
\texttt{nya} & 89.7 & 82.0 & -7.7 & 81.6 & -8.1 & 91.9 & 86.5 & -5.4 & 86.7 & -5.2 & 92.2 & 87.3 & -4.9 & 87.1 & -5.1 \\
\texttt{pcm} & 89.8 & 84.5 & -5.3 & 76.8 & -13.0 & 90.2 & 84.9 & -5.3 & 79.7 & -10.5 & 90.4 & 86.1 & -4.3 & 79.1 & -11.3 \\
\texttt{sna} & 94.9 & 89.9 & -5.0 & 93.3 & -1.6 & 95.3 & 91.4 & -3.9 & 92.4 & -2.9 & 96.3 & 93.9 & -2.4 & 93.9 & -2.4 \\
\texttt{swa} & 92.8 & 84.1 & -8.7 & 73.0 & -19.8 & 92.4 & 82.8 & -9.6 & 65.1 & -27.3 & 92.3 & 83.0 & -9.3 & 65.9 & -26.4 \\
\texttt{tsn} & 86.4 & 74.9 & -11.5 & 34.5 & -51.9 & 87.0 & 75.8 & -11.2 & 45.7 & -41.3 & 89.8 & 80.9 & -8.9 & 42.9 & -46.9 \\
\texttt{twi} & 77.9 & 65.5 & -12.4 & 52.2 & -25.7 & 80.4 & 70.9 & -9.5 & 62.3 & -18.1 & 81.4 & 72.3 & -9.1 & 63.2 & -18.2 \\
\texttt{wol} & 83.3 & 65.9 & -17.4 & 59.1 & -24.2 & 83.3 & 67.2 & -16.1 & 58.6 & -24.7 & 86.2 & 72.0 & -14.2 & 62.2 & -24.0 \\
\texttt{xho} & 88.0 & 83.2 & -4.8 & 76.7 & -11.3 & 88.0 & 83.8 & -4.2 & 76.2 & -11.8 & 90.1 & 86.5 & -3.6 & 78.5 & -11.6 \\
\texttt{yor} & 86.4 & 78.2 & -8.2 & 67.0 & -19.4 & 86.8 & 79.2 & -7.6 & 74.4 & -12.4 & 90.2 & 85.0 & -5.2 & 74.0 & -16.2 \\
\texttt{zul} & 86.4 & 83.2 & -3.2 & 69.5 & -16.9 & 89.4 & 86.1 & -3.3 & 68.8 & -20.6 & 90.1 & 87.5 & -2.6 & 67.1 & -23.0 \\
\midrule
avg & 85.5 & 76.2 & -9.3 & 70.8 & -14.7 & 86.4 & 78.0 & -8.4 & 70.9 & -15.5 & 87.5 & 79.9 & -7.6 & 72.2 & -15.3 \\
    \bottomrule
  \end{tabular}
  }
    \vspace{-3mm}
    \caption{F1 score for two varieties of hard-to-identify entities: zero-frequency entities that do not appear in the training corpus, and longer entities of four or more words.}
  \label{tab:analysis}
  \end{center}
\end{table*}

\begin{table*}[ht!]
 \begin{center}
 \resizebox{\textwidth}{!}{
   \begin{tabular}{lrrrrrrrrrrrr}
    \toprule
\multirow{2}{*}{\textbf{Language}} & \multicolumn{4}{c}{\textbf{XLM-R-large}} & \multicolumn{4}{c}{\textbf{mDeBERTaV3-base}} & \multicolumn{4}{c}{\textbf{AfroXLMR-large}} \\
    \cmidrule(lr){2-5} \cmidrule(lr){6-9} \cmidrule(lr){10-13}
& DATE & LOC & ORG & PER & DATE & LOC & ORG & PER & DATE & LOC & ORG & PER \\
    \midrule
\texttt{bam} & 90.3 & 83.2 & 80.7 & 87.1 & 90.1 & 86.4 & 79.2 & 88.4 & 92.6 & 87.7 & 82.4 & 86.1 \\
\texttt{bbj} & 87.6 & 82.9 & 79.4 & 83.6 & 79.9 & 86.4 & 72.5 & 87.2 & 85.7 & 87.0 & 75.2 & 84.7 \\
\texttt{ewe} & 91.8 & 96.8 & 85.5 & 95.9 & 91.8 & 96.4 & 88.6 & 97.1 & 92.0 & 97.8 & 85.6 & 98.6 \\
\texttt{fon} & 85.4 & 89.2 & 86.9 & 94.6 & 86.8 & 93.3 & 89.3 & 94.3 & 85.9 & 91.9 & 86.4 & 94.6 \\
\texttt{hau} & 86.8 & 90.0 & 92.5 & 98.0 & 86.4 & 89.2 & 89.1 & 98.0 & 87.4 & 91 & 92.2 & 98.2 \\
\texttt{ibo} & 84.5 & 91.6 & 83.5 & 97.7 & 85.4 & 95.6 & 82.5 & 99.1 & 87.2 & 96.5 & 73.4 & 98.8 \\
\texttt{kin} & 88.4 & 92.7 & 84.0 & 94.8 & 87.4 & 95.0 & 87.8 & 97.7 & 88.1 & 95.6 & 89.1 & 99.1 \\
\texttt{lug} & 78.2 & 93.1 & 94.2 & 95.8 & 80.2 & 95.1 & 94.3 & 96.0 & 81.7 & 93.1 & 95.1 & 97.3 \\
\texttt{mos} & 80.3 & 92.7 & 74.4 & 93.1 & 81.6 & 92.1 & 78.9 & 88.3 & 83.2 & 93.7 & 75.4 & 88.9 \\
\texttt{pcm} & 96.6 & 91.1 & 89.7 & 96.9 & 96.1 & 93.1 & 90.9 & 97.3 & 95.6 & 92.4 & 90.9 & 97.1 \\
\texttt{nya} & 89.1 & 94.1 & 94.2 & 94.4 & 89.6 & 96.7 & 96.0 & 94.9 & 89.1 & 96.2 & 94.8 & 95.6 \\
\texttt{sna} & 95.6 & 95.6 & 96.1 & 98.1 & 96.0 & 95.1 & 96.5 & 98.7 & 96.6 & 95.4 & 97.4 & 99.3 \\
\texttt{swa} & 92.2 & 97.0 & 95.2 & 98.8 & 91.5 & 96.9 & 94.6 & 98.8 & 91.5 & 97.4 & 93.7 & 98.2 \\
\texttt{tsn} & 88.1 & 88.3 & 89.1 & 97.1 & 87.8 & 90.0 & 89.0 & 97.6 & 90.5 & 94.8 & 92.2 & 98.6 \\
\texttt{twi} & 66.7 & 89.3 & 79.4 & 96.1 & 76.5 & 90.4 & 82.9 & 97.5 & 75.7 & 91.4 & 85.1 & 97.7 \\
\texttt{wol} & 80.6 & 84.9 & 87.0 & 95.9 & 80.8 & 88.2 & 88.4 & 95.0 & 82.6 & 91.9 & 88.0 & 97.0 \\
\texttt{xho} & 90.7 & 91.6 & 93.1 & 96.9 & 89.7 & 92.0 & 93.4 & 98.1 & 91.1 & 93.5 & 95.0 & 98.3 \\
\texttt{yor} & 89.6 & 94.0 & 90.3 & 93.6 & 89.6 & 92.1 & 91.4 & 94.6 & 91.3 & 95.8 & 92.5 & 96.4 \\
\texttt{zul} & 85.0 & 90.1 & 87.8 & 97.1 & 92.2 & 95.5 & 88.1 & 97.1 & 90.8 & 96.2 & 91.8 & 97.2 \\
\midrule
avg & 86.7 & 91.0 & 87.5 & 95.0 & 87.3 & 92.6 & 88.1 & 95.6 & 88.4 & 93.7 & 88.2 & 95.9 \\
    \bottomrule
  \end{tabular}
  }
    \caption{F1 score for the different entity types.}
  \label{tab:analysis2}
  \end{center}
\end{table*}

\subsection{IsiXhosa and isiZulu morphological structure}
IsiXhosa and isiZulu are agglutinative languages with a complex morphology. Each root or stem can attach a variety of affixes to form new inflections and derivations, with a variety of affixes added to root and stem morphemes to vary their meaning and convey syntactic agreement. The noun class system and the concord agreement system are the foundations of isiXhosa and isiZulu noun grammar. This section offers an overview of these two principles and their applicability to the realization of NEs. First, we briefly describe the noun class system, after which we discuss prefixing and capitalization work for both languages.

According to the Meinhoff system~\cite{melzian_1933}, nouns in African languages are classified into one of 18 numbered classes based on their prefix.  As shown in the following example, singular nouns in class 1 take the prefix um-, while associated plural nouns in class 2 take the prefix aba-.

\subsubsection{Prefix}

Even though all named entities are nouns since they designate a distinct entity, noun class designations are critical in identifying NEs. According to \citet{oosthuysen_xhosa}, the purpose of the noun class prefix is to distinguish the class to which it belongs. It shows whether the noun is singular or plural. The derivation of all significant prefixes and cordial agreements is based on this.

In isiXhosa, named entities referring to personal nouns with the prefix um- belongs to noun class 1 with noun class 2 being its plural form. Named entities such as jobs, objects and concepts belong to noun class 3, e.g. umpheki (cook) and umthwalo (burden). Lastly in isiXhosa, borrowed words from English and Afrikaans such as ibhanka (bank) and ihamire (hammer), belong to class 9. In isiZulu, noun class 1 is a singular class which uses the prefix umu-/um-. The allomorph umu- occurs when the noun stem consists of one syllable, e.g. umuntu (person) and the allomorph um- occurs when the noun stem has more than one syllable, e.g. umfana (boy). The noun class 2 is a plural class, with its singular in class 1. Noun class 2 uses
the prefix aba-/ab-, e.g. abantu (people), abafana (boys). Noun classes 1 and 2 are a personal class only containing personal nouns.

Noun class 1a is a subclass of noun class 1. This class contains personal nouns referring to family relationships, professions, proper names and personalized nouns. This class uses the prefix u- with no allomorphs, e.g. ugogo (grandmother), unesi (nurse) or uSipho (personal name). The noun class 2a is a regular plural of class 1a which uses the prefix o-, e.g. ogogo (grandmothers), onesi (nurses) or oSipho (Sipho and company).

\subsubsection{Capitalization}

Capitalization is a very common feature for a number of natural language processing tools, such as named entity recognition systems that identify people's names, and locations~\cite{De_Waal2006-dh}. Following are the four different types of the usage of capitalization in isiXhosa and isiZulu~\cite{Priatama2022-tv}: 

\begin{enumerate}
    \item Initial capitalization of words in which only the initial letter is capitalized;
    \item Mixed capitalization of words in which the initial letter of the prefix is capitalized as well as the initial letter of the main word;
    \item Internal capitalization in words which are found in the middle of a sentence where the prefix remains lower case and the first letter of the main word is capitalized.
    \item All CAPS in words that are fully capitalized. These are usually abbreviations or acronyms;
\end{enumerate}

\section{Other NER Corpus}
\label{sec:non_african_ner_data}
\autoref{tab:non_african_data_split} provides the NER corpus found online that we make use for determining the best transfer languages

\section{Error Analysis of NER}
\label{sec:error_analysis_app}
\autoref{tab:analysis} and \autoref{tab:analysis2} provides error analysis of MasakhaNER 2.0 based on performance on zero-frequency entities, long entities and distribution by named entity tags.

\section{LangRank Feature Descriptions}
\label{sec:lang_fea_descr}
The following definitions are listed here, originally from \citet{lin-etal-2019-choosing}.
\begin{description}
\item[Geographic distance ($d_{geo}$)] based on the orthodromic distance between language locations obtained from Glottolog~\citep{hammarstrom2018glottolog}. 
\item[Genetic distance ($d_{gen}$)] based on the genealogical distance of Glottolog language tree. 
\item[Inventory distance ($d_{inv}$)] based on the cosine distance between phonological feature vectors obtained from PHOIBLE database~\citep{moran2014phoible}. 
\item[Syntactic distance ($d_{syn}$)] based on cosine distance between feature vectors obtained from syntactic structures derived from WALS database~\citep{wals}. 
\item[Phonological distance ($d_{pho}$)] based on the cosine distance between phonological feature vectors obtained from WALS and Ethnologue databases~\citep{lewis2009ethnologue}. 
\item[Featural distance ($d_{fea}$)] based on the cosine distance
between feature vectors combining all 5 features mentioned above.
\item[Transfer language dataset size ($s_{tf}$)] The size of the transfer language's dataset.
\item[Target language dataset size ($s_{tg}$)] The size of the target language's dataset.
\item[Transfer over target size ratio ($sr$)] The size of the transfer language's dataset divided by the size of the target language's dataset.
\item[Entity Overlap ($eo$)] The number of unique words that overlap between the source and target languages' training datasets.
\end{description}

\begin{figure}
    \centering
    \includegraphics[width=1\linewidth]{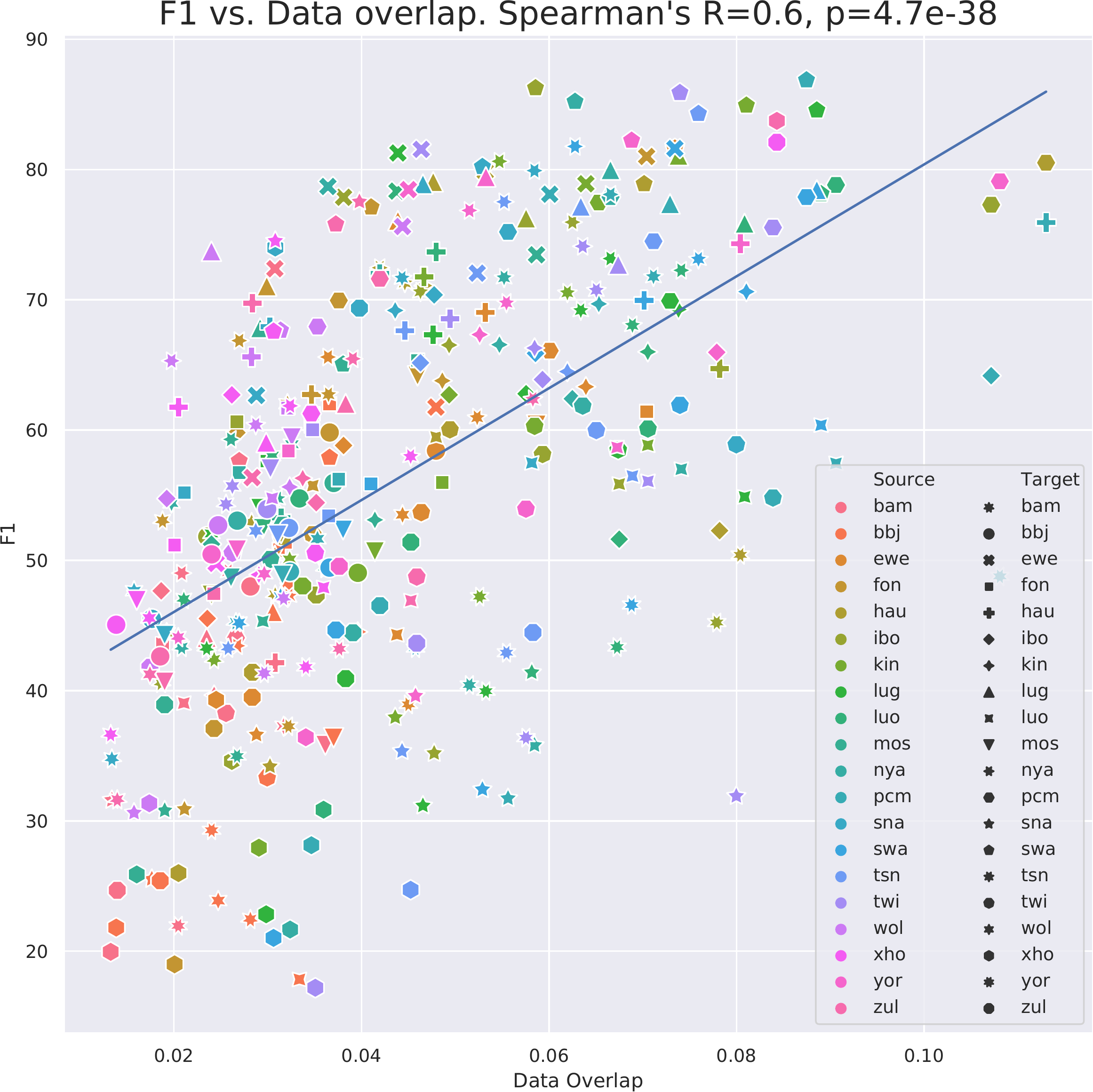}
    \caption{The correlation between the data overlap and F1 transfer performance. For source language $X$ and target language $Y$, denote the set of unique named entities (PER, ORG, LOC, DATE) by $T_X$ and $T_Y$ respectively. The overlap here was calculated as $\frac{|T_X \cap T_Y|}{|T_X| + |T_Y|}$, as in \citet{lin-etal-2019-choosing}.}
    \label{fig:appendix:overlap}
\end{figure}

\begin{figure*}[t]
    \centering
    \includegraphics[width=0.99\linewidth]{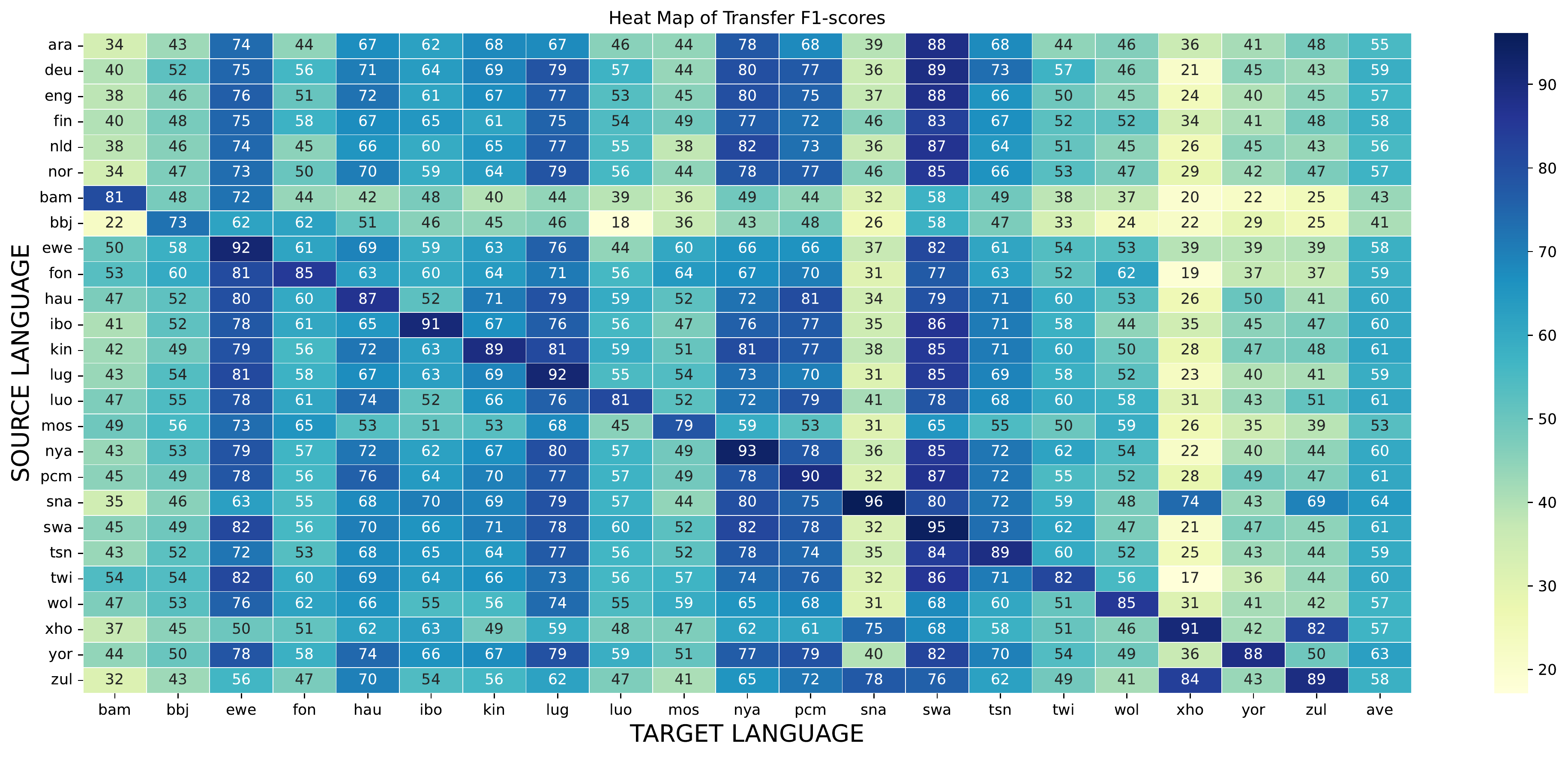}
    \vspace{-2mm}
    \caption{\textbf{Zero-shot Transfer} from several source languages to African languages in MasakhaNER 2.0.}
    \label{fig:transfer_image_0_shot_a}
\end{figure*}
\begin{figure*}[ht!]
    \centering
    \includegraphics[width=\linewidth]{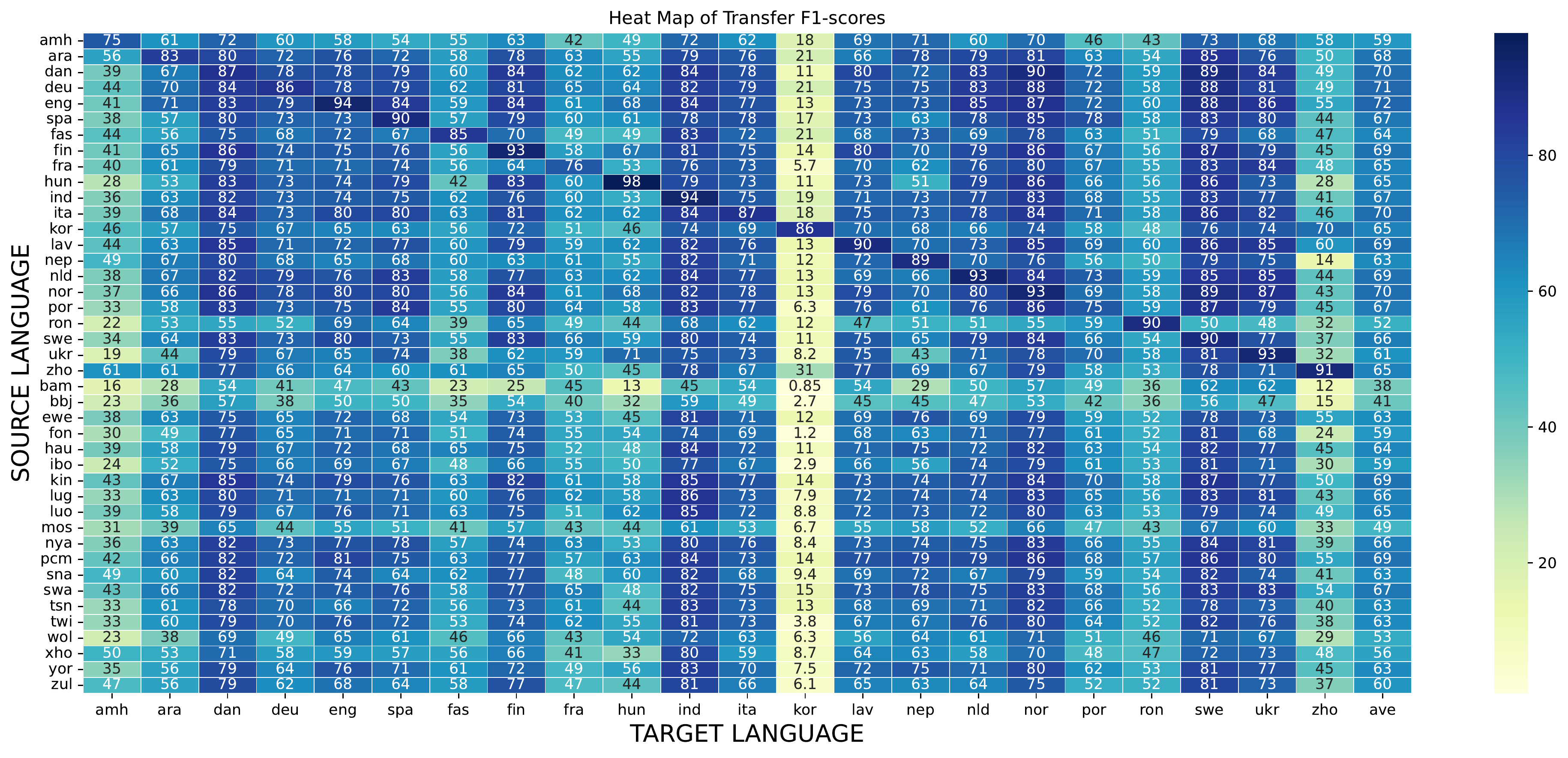}
    \vspace{-2mm}
    \caption{\textbf{Zero-shot Transfer} from several source languages to other languages not in MasakhaNER 2.0}
    \label{fig:transfer_image_0_shot_na}
\end{figure*}

\section{Overlap Results}
\label{sec:word_overlap}
In \autoref{fig:appendix:overlap}, we examine the word overlap between different languages, and how this correlates with the transfer performance. In general, these two quantities are strongly correlated (Spearman's $R = 0.6, p < 0.05$), echoing a similar result described by \citet{beukman2021analysing}. Note that the entity overlap feature used by the ranking model in the main text was calculated in a slightly different way; namely, considering \textit{all} tokens instead of just the 4 named entities and not normalizing the overlap. This case still shows a positive correlation, although it is slightly smaller with Spearman's $R=0.49$.

\section{Zero-shot Transfer}
\label{sec:zero_transfer_other_lang}

\autoref{fig:transfer_image_0_shot_a}  shows $N\times N$ transfer results to languages in MasakhaNER 2.0. We see that English is not the best transfer language in general. It is better to choose a more geographically close African language.  

\autoref{fig:transfer_image_0_shot_na}  shows $N\times N$ transfer results to languages not in MasakhaNER 2.0. We see that English appears to be the best transfer on average, which is not the case for African languages. The reason for this is that many of the non-African languages we evaluated on are from the Indo-European, similar to English.  

\section{Best Transfer Language for Other Languages}
\label{sec:best_transfer_other_lang}
\autoref{tab:best_transfer_all_languages} provides the result of the best transfer language for other languages not in MasakhaNER 2.0.

\section{Sample Efficiency Results}
\label{sec:sample_efficiency_resu}
\begin{figure*}[t]
    \centering
    \includegraphics[width=0.99\linewidth]{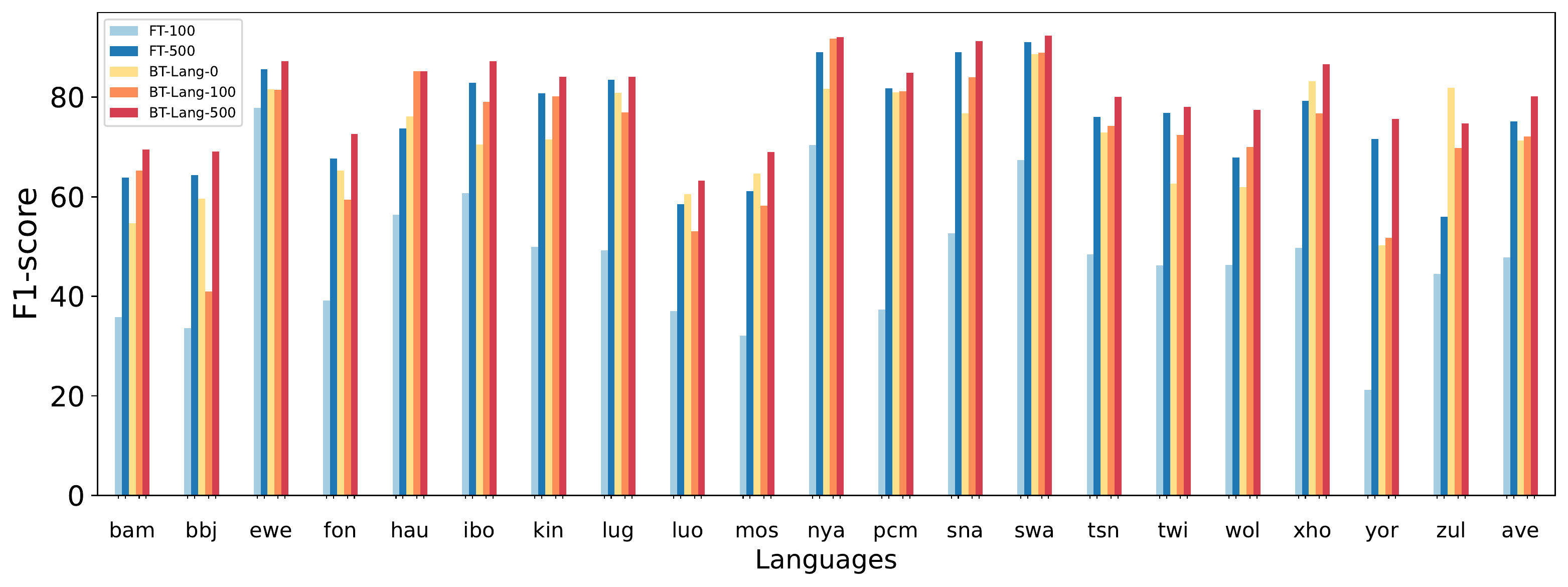}
    \vspace{-2mm}
    \caption{\textbf{Sample Efficiency Results} for 100 and 500 samples in the target language, model fine-tuned on a PLM (e.g. FT-100 -- trained on 100 samples from the target language) or fine-tuned on the best transfer language NER model (e.g. BT-Lang-0 -- trained on 0 samples from the target language or zero-shot)}
    \label{fig:transfer_image_sample_eff_all}
    
\end{figure*}

\autoref{fig:transfer_image_sample_eff_all}  shows the result of training NER models using 100 and 500 samples for each language. 

\section{Model Hyper-parameters for Reproducibility}
\label{sec:model_reproducibility}
For training NER models, we \textit{fine-tune} PLM, we make use of a maximum sequence length of 200, batch size of 16, gradient accumulation of 2, learning rate of 5e-5, and number of epochs 50.  The experiments of the large PLMs were performed on using Nvidia V100 GPU. For AfriBERTa and mBERT, we make use of Nvidia GeForce RTX-2080Ti. For evaluation, we make use of the micro-averaged F1 score.

\begin{table*}[t]
 \begin{center}
\resizebox{\textwidth}{!}{%
 \footnotesize
  \begin{tabular}{llllrrrrrr}
    \toprule
        
    
    \textbf{} & \textbf{Top-2} & \textbf{Top-2} & \textbf{Top-3 features selected} & \textbf{Target} & \textbf{Best} & \textbf{Second} & \textbf{eng} & \textbf{LangRank} & \textbf{LangRank} \\
    
    \textbf{Target} & \textbf{Transf.} & \textbf{LangRank} & \textbf{by the LangRank Model} & \textbf{Lang.} & \textbf{Transf.} & \textbf{Best} & \textbf{Tranf.} & \textbf{First} & \textbf{Second} \\
    
    \textbf{Lang.} & \textbf{Lang} & \textbf{Model} & \textbf{Lang 1; Lang 2} & \textbf{F1} & \textbf{F1} & \textbf{Transf. F1} & \textbf{F1} & \textbf{Lang F1} & \textbf{Lang F1}  \\
    \midrule
    
    \multicolumn{7}{l}{\textit{African languages}} \\ \cmidrule(lr){1-3}
    \textbf{amh} & zho, ara & pcm, luo & $(s_{tf}, s_{tg}, sr); (s_{tf}, d_{geo}, sr)$ & 75.0 & \textbf{61.0} & 55.9 & 40.6 & 42.5 & 38.6 \\ 
    \textbf{bam} & twi, fon & wol, fon & $(d_{geo}, d_{inv}, sr); (d_{geo}, sr, d_{pho})$ & 80.4 & \textbf{54.3} & 53.0 & 38.4 & 47.1 & 53.0\\ 
    
    \textbf{bbj} & fon, ewe & twi, ewe & $(s_{tf}, d_{syn}, d_{geo}); (s_{tf}, d_{geo}, sr)$ & 72.9 & \textbf{59.8} & 58.4 & 45.8 & 53.9 & 58.4\\
    
    \textbf{ewe} & swa, twi & pcm, swa & $(d_{geo}, s_{tf}, sr); (eo, d_{geo}, s_{tf})$ & 91.7 & \textbf{81.6} & 81.5 & 76.4 & 78.1 & \textbf{81.6} \\ 
    
    \textbf{fon} & mos, bbj & yor, ewe & $(d_{geo}, d_{syn}, sr); (s_{tf}, d_{geo}, d_{gen})$ & 84.9 & \textbf{65.4} & 62.0 & 50.6 & 58.4 & 61.4\\
    
    \textbf{hau} & pcm, yor & yor, swa & $(d_{geo}, sr, eo); (eo, sr, s_{tf})$ & 86.9 & 75.9 & \textbf{74.3} & 72.4 & 74.3 & 70.0 \\ 
    
    \textbf{ibo} & sna, yor & pcm, kin & $(eo, d_{geo}, s_{tf}); (d_{geo}, sr, eo)$ & 91.0 & \textbf{70.4} & 66.0 & 61.4 & 64.2 & 62.7 \\ 
    
    \textbf{kin} & hau, swa & sna, yor & $(eo, d_{geo}, s_{tf}); (eo, s_{tf}, sr)$ & 89.5 & \textbf{71.1} & 70.6 & 67.4 & 69.2 & 67.3\\
    
    \textbf{lug} & kin, nya & luo, zul & $(d_{geo}, sr, eo); (d_{syn}, d_{geo}, sr)$ & 91.5 & \textbf{81.1} & 80.0 & 76.5 & 75.9 & 62.0 \\ 
    
    \textbf{luo} & swa, hau & lug, sna & $(d_{geo}, sr, eo); (d_{geo}, eo, sr)$ & 81.2 & \textbf{60.4} & 59.5 & 53.4  & 54.9 & 57.5 \\
    
   \textbf{mos} & fon, ewe & yor, fon & $(d_{geo}, d_{inv}, sr); (d_{geo}, s_{tf}, sr)$ & 78.9 & \textbf{64.2} & 60.4 & 45.4&  50.8 & \textbf{64.2}\\
   
    \textbf{nya} & swa, nld & zul, sna & $(eo, d_{geo}, sr); (d_{geo}, eo, d_{syn})$ & 93.5 & \textbf{81.8} & 81.7 & 80.1 & 65.5 & 79.9 \\
    
    \textbf{pcm} & hau, yor & eng, yor & $(eo, d_{gen}, d_{syn}); (eo, d_{geo}, sr)$ & 89.9 & \textbf{80.5} & 79.1 & 75.5 & 75.5 & 79.1 \\ 
    
    \textbf{sna} & zul, xho & swa, zul & $(eo, sr, s_{tf}); (d_{geo}, sr, eo)$ & 96.0 & \textbf{77.5} & 74.5 & 37.1 & 32.4 & 77.5\\
    
    \textbf{swa} & deu, ara & ita, nld & $(sr, d_{inv}, eo); (eo, s_{tf}, sr)$ & 94.6 & \textbf{88.7} & 88.1 & 87.9 & 84.5 & 86.6 \\ 
    
    \textbf{tsn} & deu, swa & swa, nya & $(eo, d_{inv}, s_{tf}); (d_{inv}, d_{geo}, d_{gen})$ & 88.7 & \textbf{73.3} & 73.1 & 65.8 & 73.1 & 71.7 \\ 
    
    \textbf{twi} & swa, nya & swa, ewe & $(eo, s_{tf}, d_{geo}); (d_{geo}, s_{tf}, sr)$ & 82.0 & 61.0 & \textbf{61.9} & 49.5 & \textbf{61.9} & 53.7\\
    
    \textbf{wol} & fon, mos & fon, yor & $(d_{geo}, sr, s_{tf}); (sr, d_{geo}, d_{syn})$ & 85.2 & \textbf{62.0} & 58.9 & 44.8 & \textbf{62.0} & 49.0 \\ 
    
    \textbf{xho} & zul, sna & zul, pcm & $(eo, d_{geo}, d_{gen}); (eo, s_{tf}, d_{inv})$ & 90.8 & \textbf{83.7} & 74.0 & 24.5 & 83.7 & 28.1 \\
    
    \textbf{yor} & hau, pcm & fon, pcm & $(d_{geo}, d_{inv}, d_{syn}); (eo, d_{geo}, d_{inv})$ & 88.3 & \textbf{50.3} & 48.8 & 40.1  & 37.3 & 48.8\\
    
    \textbf{zul} & xho, sna & xho, sna & $(eo, d_{gen}, d_{geo}); (d_{syn}, sr, d_{geo})$ & 88.6 & \textbf{82.1} & 69.4 & 44.7 & \textbf{82.1} & 69.4 \\

    \midrule
    \multicolumn{7}{l}{\textit{Non-African languages}} \\ \cmidrule(lr){1-3}
    \textbf{ara} & eng, deu & fas, pcm & $(eo, d_{inv}, d_{syn}); (d_{syn}, sr, d_{inv})$ & 82.8 & \textbf{71.5} & 69.9 & \textbf{71.5} & 55.7 & 57.9\\ 
    \textbf{dan} & nor, fin & swe, nor & $(eo, d_{gen}, d_{geo}); (eo, d_{geo}, d_{syn})$ & 87.1 & \textbf{86.3} & 85.6 & 83.1 & 82.8 & 86.3\\
    
    \textbf{deu} & nld, eng & dan, nld & $(d_{geo}, eo, s_{tf}, d_{syn}); (eo, d_{syn}, d_{geo})$ & 86.5 & \textbf{79.3} & 78.8 & 78.8 & 79.3 & 79.3\\
    
    \textbf{eng} & pcm, swe & nld, pcm & $(eo, d_{geo}, d_{syn}); (eo, d_{gen} d_{pho})$ & 93.5 & 81.3 & 79.7 & \textbf{93.5} & 76.0 & 81.3\\ 
    
    \textbf{fas} & hau, pcm & ara, eng & $(d_{syn}, d_{inv}, eo); (d_{syn}, d_{geo}, s_{tf})$ & 84.8 & \textbf{64.8} & 63.4 & 59.3  & 57.9 & 59.2\\ 
    
    \textbf{fin} & dan, eng & deu, eng & $(eo, s_{tf},d_{geo}); (d_{syn}, d_{geo}, eo)$ & 93.4 & \textbf{83.7} & 83.6 & 83.6 & 80.8 & 83.6\\ 
    
    \textbf{fra} & swe, swa & nld, deu & $(eo, d_{syn}, d_{geo}); (d_{geo}, eo, sr)$ & 75.5 & \textbf{66.3} & 65.4 & 60.6 & 63.3 & 64.9\\
    
    \textbf{hun} & ukr, eng & deu, ron & $(d_{geo}, d_{syn}, eo); (d_{geo}, eo, d_{syn})$ & 98.0 & \textbf{70.7} & 68.4 & 68.4 & 63.6 & 43.8\\ 
    
    \textbf{ind} & lug, luo  & zho, nld & $(s_{tg}, s_{tf}, sr); (d_{syn}, s_{tf}, eo)$ & 93.7 & \textbf{85.9} & 85.2 & 83.9 & 78.6 & 84.1\\
    
   \textbf{ita} & deu, spa & nld, eng & $(d_{syn}, eo, d_{geo}); (eo, d_{syn}, d_{geo})$ &86.7& \textbf{79.1} & 78.2 & 77.0 & 77.1 & 77.1\\
   
    \textbf{kor} & zho, ind & ara, nep & $(sr, s_{tf}, d_{syn}); (d_{inv}, d_{syn}, s_{tf})$  & 85.7 & \textbf{31.1} & 21.5 & 12.7 & 21.3 & 11.9\\
    
    \textbf{lav} & fin, dan & eng, nld & $(s_{tf}, d_{syn}, sr); (s_{tf}, d_{syn}, d_{geo})$ & 89.7 & \textbf{80.4} & 80.1 & 73.5 & 73.5 & 69.5 \\ 
    
    \textbf{nep} & pcm, swa & kor, zho & $(d_{syn}, s_{tf}, d_{pho}); (s_{tf}, sr, d_{geo})$  & 89.5 & \textbf{79.0} & 77.7 & 73.4 & 68.2 & 68.5\\
    
    \textbf{nld} & eng, deu & eng, nor & $(eo, d_{geo}, d_{syn}); (eo,d_{geo}, s_{tf})$ & 93.4 & \textbf{85.4} & 83.7 & 85.4 & \textbf{85.4} & 79.9\\  
    
    \textbf{nor} & dan, deu & dan, eng & $(eo, d_{geo}, s_{tf}); (eo, d_{geo}, sr)$ & 92.5 & \textbf{89.8} & 87.8 & 87.3 & 89.8 & 87.2 \\  
    
    \textbf{por} & es, nld & spa, eng & $(eo, d_{syn}, d_{gen}); (eo, d_{syn}, d_{geo})$  & 75.0 & \textbf{77.8} & 73.5 & 72.0 & 77.8 & 72.0 \\
    
    \textbf{ron} & lav, eng & eng, ita & $(eo, d_{syn}, d_{geo}); (eo, d_{geo}, d_{syn})$& 89.6 & \textbf{59.6} & 59.5 & 59.5 & 59.5 & 57.8\\ 
    
    \textbf{spa} & eng, por & por, lav & $(eo, d_{geo}, d_{syn}); (d_{syn}, eo, d_{geo})$ & 89.6 & \textbf{83.9} & 83.6 & \textbf{83.9} & 83.6 & 77.3\\
    
    \textbf{swe} & dan, nor & nor, nld & $(eo, d_{syn}, d_{geo}); (d_{syn}, d_{geo}, eo)$ & 90.3 & \textbf{89.4} & 89.1 & 88.1 & 89.3 & 85.2 \\
    
    \textbf{ukr} & nor, eng & deu, eng & $(d_{geo}, d_{syn}, sr); (d_{syn}, d_{geo}, s_{tf})$  & 92.6 & \textbf{87.2} & 85.6 & 85.6 & 81.5 & 85.6\\
    
    \textbf{zho} & lav, amh & pcm, deu & $(d_{syn}, s_{tf}, s_{geo}); (d_{syn}, s_{tf}, d_{pho})$ & 91.4 & \textbf{60.2} & 58.3 & 54.7 & 54.7 & 48.9 \\ 
    \midrule
    AVG & -- & & & 87.7 & 73.3 & 71.2 & 64.6 & 67.3 & 66.2\\ 

    \bottomrule
  \end{tabular}
  }
  \vspace{-3mm}
  \caption{\textbf{Best Transfer Language for NER.} 
  The ranking model features are based on the definitions in \cite{lin-etal-2019-choosing} like: geographic distance ($d_{geo}$), genetic distance ($d_{gen}$), inventory distance ($d_{inv}$), syntactic distance ($d_{syn}$), phonological distance ($d_{pho}$), transfer language dataset size ($s_{tf}$), target language dataset size($s_{tg}$),  transfer over target size ratio ($sr$), and entity overlap ($eo$). }
  \label{tab:best_transfer_all_languages}
  \end{center}
\end{table*}

\end{document}